\PassOptionsToPackage{table,xcdraw}{xcolor}
\documentclass[acmlarge,nonacm]{acmart}
\setcopyright{none}
\acmDOI{10.1145/3836772}

\usepackage{booktabs}
\usepackage{multirow}
\usepackage{graphicx}
\usepackage{colortbl}
\usepackage{array}
\AtBeginDocument{\ifdefined\symbfup\let\mathbf\symbfup\fi}

\newcolumntype{C}[1]{>{\centering\arraybackslash}p{#1}}

\newcommand{\rankone}[1]{\cellcolor[HTML]{F4B084}\makebox[\linewidth][c]{\textcolor[HTML]{9C0006}{\textbf{#1}}}}
\newcommand{\ranktwo}[1]{\cellcolor[HTML]{FCD5B4}\makebox[\linewidth][c]{\textcolor[HTML]{DC2828}{\textbf{#1}}}}
\newcommand{\rankthree}[1]{\cellcolor[HTML]{FFE5CC}\makebox[\linewidth][c]{\textcolor[HTML]{FF7878}{\textbf{#1}}}}

\AtBeginDocument{%
  }

\begin{document}

\title{Geometry-Constrained Bidirectional Point Cloud Registration for Thin, Sheet-Like Heritage Artifacts}


\author{Yuezhe Zhang}
\orcid{0009-0000-2790-2512}
\email{yuezhezhang@mail.nwpu.edu.cn}
\affiliation{%
  \institution{Northwestern Polytechnical University}
  \city{Xi'an}
  \state{Shaanxi}
  \country{China}
}

\author{Lei Wei}
\authornote{Corresponding author.}
\orcid{0000-0002-0958-4597}
\email{l.wei@mail.nwpu.edu.cn}
\affiliation{%
  \institution{Xi'an International University}
  \city{Xi'an}
  \state{Shaanxi}
  \country{China}
}

\author{Jingnan Du}
\email{dujingnan@nwpu.edu.cn}
\orcid{0000-0002-5623-2367}
\affiliation{%
  \institution{Northwestern Polytechnical University}
  \city{Xi'an}
  \state{Shaanxi}
  \country{China}
}

\author{Shuai Wan}
\orcid{0000-0001-8617-149X}
\email{swan@nwpu.edu.cn}
\affiliation{%
  \institution{Northwestern Polytechnical University}
  \city{Xi'an}
  \state{Shaanxi}
  \country{China}
}

\renewcommand{\shortauthors}{Yuezhe Zhang, Lei Wei, Jingnan Du, and Shuai Wan}

\begin{abstract}

Non-contact three-dimensional reconstruction of thin, sheet-like heritage artifacts poses significant geometric and registration challenges. Due to their fragility, these artifacts cannot be suspended or equipped with artificial markers, necessitating independent acquisition of their front and back surfaces. Subsequent registration proves difficult due to the limited number of shared geometric features and the scarcity of explicit physical constraints, which may result in rotational ambiguity, instability, and structural collapse during iterative optimization. To address these challenges, we propose a geometry-constrained bidirectional point cloud registration method specifically tailored for thin, sheet-like heritage artifacts. The method integrates semantic-guided preprocessing, Principal Component Analysis (PCA)-based geometric normalization, and a thickness-aware registration strategy. The estimated physical thickness is incorporated as a geometric constraint to preserve structural integrity during registration. Rotational ambiguity is resolved by evaluating a finite set of global rotation hypotheses, each refined using the point-to-plane Iterative Closest Point (ICP) algorithm, with the optimal transformation selected via a geometry-aware fitness criterion consistent with the thickness scale. Experimental results show that the proposed method achieves competitive or improved performance in most cases, particularly in projected area consistency and physically plausible front–back alignment. In addition, the thickness-aware constraint and rotation hypothesis evaluation reduce the risk of degenerate configurations in which the two surfaces are incorrectly flipped while still yielding deceptively acceptable numerical scores, supporting reliable non-contact digitization of delicate and thin heritage artifacts. Implementation details are available at \url{https://zyz-nwpu.github.io/GCBPCR/}.

\end{abstract}



\maketitle

\begin{center}
\footnotesize
\textcopyright\ 2026 Copyright held by the owner/author(s).\\
\emph{Journal on Computing and Cultural Heritage} (2026).\\
\url{https://doi.org/10.1145/3836772}
\end{center}

\section{Introduction}

The digital preservation of cultural heritage artifacts has become an important component of modern conservation practice, enabling systematic documentation, structural analysis, virtual restoration, and long-term digital archiving~\cite{remondino2011heritage}. Accurate three-dimensional (3D) models allow fragile historical objects to be studied and preserved while minimizing physical handling. Early large-scale digitization efforts relied on active range sensing technologies such as laser scanning~\cite{levoy2000digital}. Structured-light scanning systems were later introduced to provide dense and accurate surface measurements for smaller artifacts and museum collections~\cite{salvi2007review}. Due to their ability to capture precise geometric details, these systems have been widely adopted in cultural heritage digitization workflows~\cite{guidi2014three}. However, active scanning methods typically require specialized hardware, controlled acquisition environments, and complex calibration procedures.

With the rapid development of computer vision, image-based reconstruction has become a flexible solution for heritage digitization. Structure-from-Motion (SfM) estimates camera poses and sparse geometry from unordered images~\cite{snavely2006photo}, while later work improved the robustness and scalability of SfM pipelines for large datasets~\cite{schonberger2016sfm}. Efficient incremental systems further enabled large-scale reconstruction from thousands of images~\cite{wu2013towards}. Multi-View Stereo (MVS) methods reconstruct dense surface geometry from calibrated images~\cite{furukawa2010accurate}. Subsequent improvements introduced more robust view selection and pixelwise optimization strategies for dense reconstruction~\cite{schoenberger2016mvs}. These SfM–MVS pipelines have become a standard workflow for image-based 3D reconstruction and are widely used for digitizing archaeological sites and museum artifacts~\cite{koutsoudis2014multi}. Close-range photogrammetry is also widely used in heritage documentation due to its portability and relatively low cost~\cite{luhmann2013close}. Although these techniques are mature, fragile artifacts often require non-contact acquisition from multiple orientations, making point cloud registration necessary to recover complete geometry.

Among these objects, thin sheet-like artifacts such as bronze fragments, ceramic shards, and thin wooden panels present particular challenges for 3D reconstruction. Their extremely small thickness relative to lateral dimensions results in near-planar geometry with limited structural depth. Due to conservation constraints, these fragile artifacts cannot be suspended or equipped with artificial markers during acquisition and must instead be placed on a supporting surface. Consequently, the front and back sides are captured separately under non-contact conditions consistent with non-invasive heritage digitization principles~\cite{guidi2014three}. Because the two sides are reconstructed independently, their point clouds lie in different local coordinate systems and require accurate rigid registration to recover the complete geometry.

Estimating this transformation remains challenging for thin-sheet artifacts due to their near-planar geometry and limited thickness variation, which reduce geometric saliency and weaken feature-based registration methods. Classical optimization approaches such as Iterative Closest Point (ICP) estimate rigid transformations by minimizing distances between corresponding points~\cite{besl1992method}, but they do not incorporate thickness-aware constraints and may lead to degenerate solutions. Marker-based registration is unsuitable because attaching artificial markers violates the non-invasive principles of cultural heritage digitization~\cite{remondino2011heritage}. Learning-based registration methods have also been proposed, but they typically require large annotated datasets and do not explicitly enforce geometry-aware thickness constraints~\cite{qin2022geotransformer}. Therefore, robust registration frameworks for thin-sheet heritage artifacts acquired under non-contact conditions remain limited.

To address these limitations, this work proposes a geometry-constrained bidirectional point cloud registration method specifically designed for thin, sheet-like heritage artifacts. The proposed method integrates semantic-guided preprocessing, Principal Component Analysis (PCA)-based geometric normalization, and thickness-aware registration.

The main contributions of this work are summarized as follows:

\textbf{Thickness-aware geometric constraint modeling:}
We propose a robust percentile-based thickness estimation method within a PCA-normalized coordinate system and explicitly incorporate the estimated physical thickness into the registration process as a geometric prior, effectively preventing non-physical structural collapse during iterative optimization.

\textbf{Global rotation hypothesis evaluation for ambiguity resolution:}
We introduce a finite rotation hypothesis strategy derived from principal-axis sign permutations to systematically resolve orientation ambiguity in symmetric thin structures. Each hypothesis is refined via point-to-plane ICP, and the optimal transformation is selected using a geometry-consistent fitness criterion.

\textbf{Semantic-guided structural purification:}
We integrate image-level semantic masking with three-dimensional spatial clustering to obtain structurally clean point clouds prior to registration, enabling non-contact background removal without the need for physical markers.

\textbf{A practical non-invasive method for thin-sheet heritage digitization:}
The proposed geometry-constrained bidirectional registration pipeline provides a stable and physically consistent approach for front-back reconstruction of fragile thin-sheet artifacts under non-contact acquisition conditions.

This paper is organized as follows. Section 2 reviews related work on point cloud registration and challenges associated with thin structures. Section 3 presents the proposed geometry-constrained registration method, including semantic-guided purification, PCA-based canonical registration, thickness estimation, rotation hypothesis search, and ICP refinement under physical constraints. Section 4 reports experimental results and ablation studies on representative thin-sheet heritage objects. Finally, Section 5 concludes the paper and discusses limitations and future work.

\section{Related Work}

\subsection{Point Cloud Registration}

Point cloud registration aims to estimate the rigid transformation that aligns multiple three-dimensional point clouds into a common coordinate system. It is a fundamental problem in computer vision, robotics, and 3D perception, playing a critical role in applications such as 3D reconstruction, SLAM, autonomous driving, and robotic manipulation. Over the past several decades, extensive research has been conducted on point cloud registration, leading to a wide variety of algorithms that can generally be categorized into local iterative methods, global registration approaches, handcrafted feature-based techniques, and more recently learning-based methods.

One of the earliest and most widely used registration algorithms is the Iterative Closest Point (ICP) method introduced by Besl and McKay \cite{besl1992method}. ICP iteratively establishes correspondences between two point clouds and estimates the rigid transformation that minimizes the Euclidean distance between matched points. Due to its conceptual simplicity and practical effectiveness, ICP has become a fundamental component in many 3D processing pipelines. Numerous variants have been proposed to improve its convergence speed and robustness. Chen and Medioni \cite{chen1992object} proposed the point-to-plane ICP formulation, which minimizes the distance between points and their corresponding tangent planes, leading to significantly faster convergence. Segal et al. \cite{Segal2009GICP} further proposed the Generalized ICP framework, which unifies point-to-point and point-to-plane formulations under a probabilistic model and improves robustness in noisy environments. Rusinkiewicz and Levoy \cite{rusinkiewicz2001efficient} also presented an efficient variant of ICP that improves performance through optimized correspondence selection and error metrics.

Despite their effectiveness, ICP-based methods generally require a good initial alignment and sufficient overlap between point clouds. To overcome this limitation, global registration methods have been proposed to estimate transformations without prior alignment. Aiger et al. \cite{aiger2008fpcs} introduced the Four-Point Congruent Sets (4PCS) algorithm, which identifies congruent geometric configurations between two point clouds by matching sets of four coplanar points. This approach enables robust registration even under large transformations and limited overlap. Mellado et al. \cite{mellado2014super} later proposed Super4PCS, which significantly improves the efficiency of the original algorithm through smart indexing and geometric filtering, enabling near-linear computational complexity.

Another important class of methods relies on local geometric descriptors to establish correspondences between point clouds. These feature-based approaches compute invariant descriptors that capture the geometric properties of local neighborhoods. Rusu et al. \cite{rusu2009fast} proposed the Fast Point Feature Histogram (FPFH), which efficiently encodes angular relationships between neighboring points and has been widely used for coarse alignment in 3D registration tasks. Tombari et al. \cite{tombari2010unique} introduced the Unique Shape Context (USC) descriptor, which improves robustness for object recognition and registration tasks. Salti et al. \cite{salti2014shot} proposed the SHOT descriptor, which combines histograms of surface orientations with spatial partitioning to produce highly discriminative local descriptors.

In recent years, deep learning has emerged as a powerful paradigm for point cloud registration. Instead of relying on handcrafted descriptors, learning-based methods aim to learn feature representations directly from raw point cloud data. Aoki et al. \cite{aoki2019pointnetlk} introduced PointNetLK, which integrates the PointNet architecture with the classical Lucas–Kanade optimization framework for rigid point cloud alignment. Choy et al. \cite{choy2019fully} proposed Fully Convolutional Geometric Features (FCGF), which learns dense local features for robust correspondence matching across point clouds. Bai et al. \cite{bai2020d3feat} further proposed D3Feat, which jointly learns keypoint detection and descriptor extraction for 3D point clouds.

More recent work has focused on improving correspondence estimation through attention mechanisms and transformer-based architectures. Huang et al. \cite{huang2021predator} proposed Predator, which introduces overlap-aware attention mechanisms to improve feature matching under low-overlap scenarios. Qin et al. \cite{qin2022geotransformer} introduced GeoTransformer, which leverages geometric attention to capture long-range spatial relationships and significantly improves robustness under large viewpoint changes.

Although these methods have achieved remarkable progress, most existing registration algorithms still rely on assumptions such as sufficient overlap and distinctive geometric features between point clouds. In practical scenarios involving thin or near-planar structures, these assumptions are often violated. In such cases, the lack of distinctive geometric features can lead to ambiguous correspondences and unstable optimization, motivating the development of more specialized registration frameworks.

\subsection{Registration of Thin and Low-Overlap Structures}

Although substantial progress has been made in point cloud registration, most existing methods implicitly assume that the target objects exhibit sufficiently rich three-dimensional geometry and overlapping surface regions. These assumptions are often violated when dealing with thin, sheet-like artifacts or near-planar structures, where geometric degeneracy significantly complicates the registration process.

A fundamental challenge arises from the limited geometric variation present in thin or planar surfaces. When the majority of points lie close to a two-dimensional manifold, the covariance structure of the point cloud becomes highly anisotropic, and the smallest principal component often corresponds to the thickness direction of the surface. Under such conditions, classical registration algorithms such as ICP may suffer from rank-deficient optimization problems, leading to unstable convergence or ambiguous transformations \cite{rusinkiewicz2001efficient}. Similar degeneracy issues have been discussed in geometric modeling and surface alignment literature, where planar or symmetric structures reduce the observability of rigid transformations \cite{pomerleau2013comparing}.

Another difficulty is the scarcity of distinctive local geometric features. Feature-based registration methods rely on discriminative descriptors computed from local neighborhoods, such as FPFH \cite{rusu2009fast} or SHOT \cite{salti2014shot}. However, when applied to smooth or low-curvature surfaces, these descriptors often become less distinctive due to the lack of geometric variation, resulting in unreliable correspondences. This limitation has been widely reported in empirical evaluations of registration algorithms on structured environments containing large planar regions \cite{pomerleau2015review}.

More recent learning-based approaches have attempted to address challenging scenarios such as partial overlap and large viewpoint changes. For example, Predator \cite{huang2021predator} introduces an overlap-aware attention mechanism to improve correspondence estimation in low-overlap situations, while GeoTransformer \cite{qin2022geotransformer} leverages geometric attention to capture long-range spatial relationships between points. Despite their strong performance in many benchmark datasets, these methods still assume the existence of at least partial overlapping regions between point clouds. In scenarios involving double-sided acquisition of thin objects, the front and back surfaces may share little or no overlapping geometry, making correspondence-based matching fundamentally ill-posed.

Furthermore, thin sheet-like objects introduce additional physical constraints that are rarely considered in conventional registration frameworks. In practical digitization workflows, the two sides of an artifact are often captured independently due to conservation constraints, and the resulting point clouds correspond to opposite surfaces separated by a small estimated physical thickness. Without incorporating such geometric priors, standard registration algorithms may produce physically implausible solutions, including surface interpenetration or collapse along the thickness direction.

These limitations highlight the need for specialized registration strategies that explicitly consider the geometric characteristics of thin structures and the acquisition conditions of double-sided scanning. In particular, incorporating structural priors such as thickness constraints or canonical geometric normalization can help stabilize the alignment process and reduce ambiguity in under-constrained configurations.

\section{Proposed Method}

\begin{figure}[t]
\centering
\includegraphics[width=\linewidth]{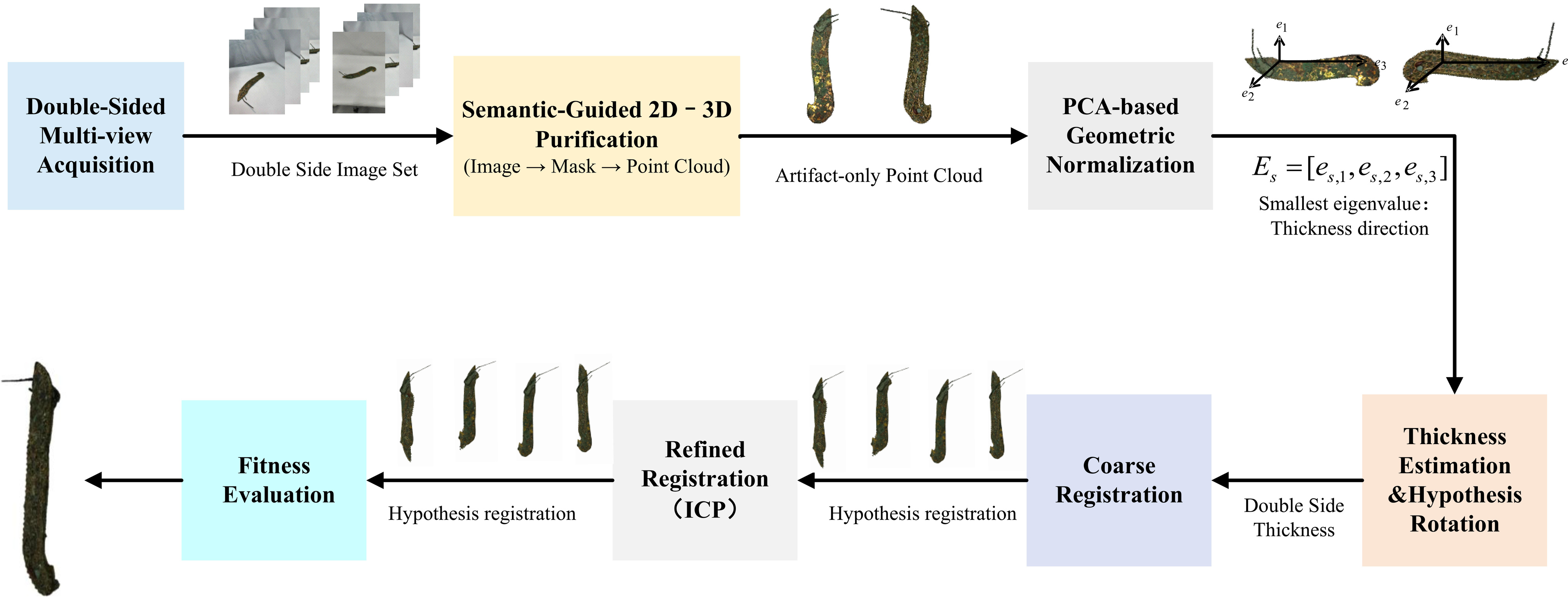}
\caption{Overview of the proposed geometry-constrained bidirectional point cloud registration pipeline for thin sheet-like artifacts. The workflow begins with independent front- and back-side multi-view reconstruction using an SfM--MVS pipeline. Semantic-guided purification is then applied to remove background structures and reconstruction artifacts, producing artifact-only point clouds for both sides. Next, PCA-based canonical normalization aligns each point cloud to its intrinsic geometric axes and enables robust estimation of the artifact thickness. Based on this canonical representation, a geometry-constrained registration stage evaluates a finite set of rotation hypotheses and refines each candidate using point-to-plane ICP initialized with a thickness-aware offset. The optimal transformation is selected using a thickness-aware fitness criterion, and the aligned point clouds are finally merged to obtain the complete artifact model.}
\label{fig:pipeline}
\Description{A flowchart of the complete registration pipeline. Double-sided image acquisition and semantic-guided purification produce separate artifact-only point clouds for the front and back surfaces. PCA aligns each surface with three principal axes, with the smallest-eigenvalue axis representing thickness. Thickness estimation generates several rotation hypotheses, which pass through coarse registration, point-to-plane ICP refinement, and fitness evaluation. The highest-scoring hypothesis produces the final merged artifact model.}
\end{figure}

\begin{figure}[t]
  \centering
  \includegraphics[width=\linewidth]{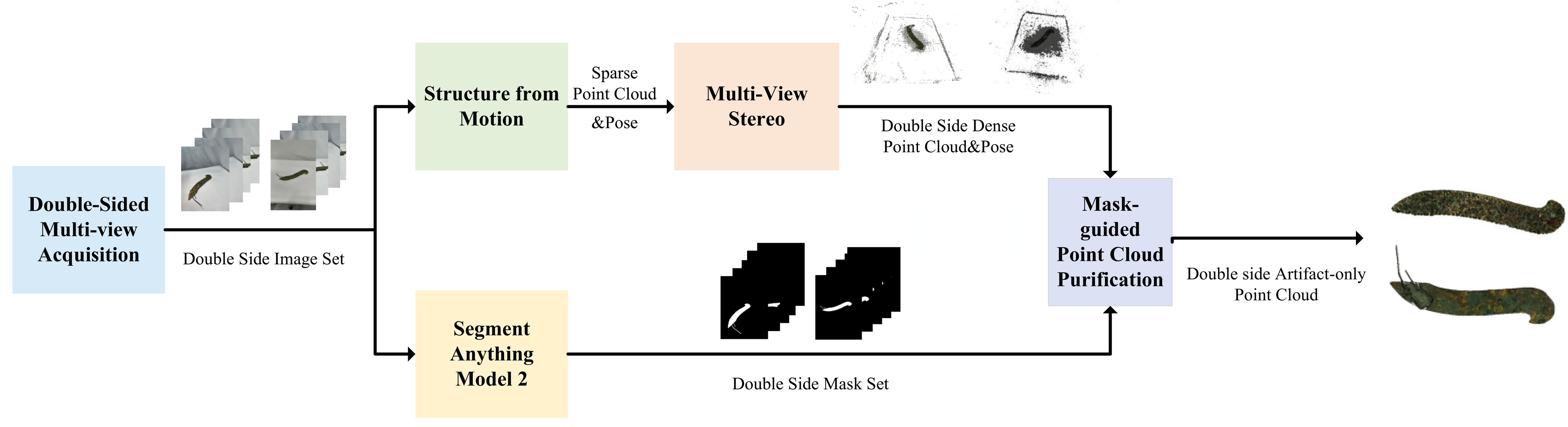}
  \caption{Artifact-aware multi-view reconstruction and purification pipeline. The front and back surfaces are digitized through double-sided independent multi-view acquisition. The captured image sets are processed using an SfM--MVS reconstruction framework to generate dense point clouds. Semantic masks are predicted for each image and integrated with 3D structural filtering to refine the reconstructed data, yielding artifact-only point clouds for subsequent registration.}
  \Description{A two-branch purification workflow beginning with front-side and back-side image sets. The upper branch applies Structure from Motion and Multi-View Stereo to estimate camera poses and dense point clouds. The lower branch applies Segment Anything Model 2 to generate binary artifact masks. Camera information, dense point clouds, and masks converge at mask-guided point-cloud purification, producing clean and separate point clouds for the two artifact surfaces.}
  \label{fig:framework}
\end{figure}

The proposed method consists of three main stages, as illustrated in Fig.~\ref{fig:pipeline}. First, a semantic-guided 2D--3D purification process is performed to remove background structures and reconstruction artifacts and obtain artifact-only point clouds (Section~3.1). As further detailed in Fig.~\ref{fig:framework}, this stage expands the artifact-aware multi-view reconstruction and purification part of Fig.~1, including double-sided image acquisition, SfM--MVS reconstruction, semantic masking, and mask-guided point cloud purification. Next, PCA-based canonical normalization is performed to establish intrinsic geometric axes and estimate the artifact thickness (Section~3.2). Finally, a geometry-constrained bidirectional registration strategy aligns the front and back surfaces by combining rotation hypothesis evaluation with point-to-plane ICP refinement under a thickness-aware constraint (Section~3.3).

\subsection{Semantic-Guided 2D--3D Purification}

Following the artifact-aware pipeline in Fig.~1, we treat the front and back surfaces of thin, sheet-like heritage artifacts as two independent acquisition-and-reconstruction sessions, and we purify each session in its own coordinate system before any cross-side registration is attempted. This separation is essential because the supporting plane is present in both sessions but appears differently under viewpoint and illumination changes, and the reconstructed shells may contain side-specific background adhesion and outliers. Accordingly, we define two disjoint multi-view datasets and process them with the same purification operator, instantiated once for the front side and once for the back side.

For the front side, let the captured front-side image sequence be
\begin{equation}
\mathcal{I}_f = \{I_f^k\}_{k=1}^{N_f},
\end{equation}
where each image is defined as
\begin{equation}
I_f^k : \Omega_f^k \subset \mathbb{R}^2 \rightarrow \mathbb{R}^3,
\end{equation}
with \(\Omega_f^k\) denoting the pixel domain and \(\mathbb{R}^3\) representing RGB color space.

To suppress background interference introduced by the supporting plane,
we employ the Segment Anything Model 2 (SAM2) \cite{sam2} to obtain artifact-aware
semantic masks. Specifically, foreground and background prompts are manually provided
on the first image of the sequence, and SAM2 propagates the segmentation
to the remaining frames, producing a binary mask for each image.

Let the resulting mask set be
\begin{equation}
\mathcal{M}_f=\{M_f^k\}_{k=1}^{N_f}, \qquad
M_f^k:\Omega_f^k\rightarrow\{0,1\},
\end{equation}
where \(1\) denotes artifact pixels and \(0\) denotes background.

The front-side image sequence is processed using a standard SfM--MVS pipeline, producing a dense point cloud representing the front surface of the artifact

\begin{equation}
\mathcal{P}_f = \{\mathbf{X}_i^{(f)} \in \mathbb{R}^3\}_{i=1}^{M_f},
\end{equation}
together with the corresponding camera parameters
\begin{equation}
\{\mathbf{K}_f^k, \mathbf{R}_f^k, \mathbf{t}_f^k\}_{k=1}^{N_f},
\end{equation}
where \(N_f\) denotes the number of front-side images.
For each view \(k\), \(\mathbf{K}_f^k \in \mathbb{R}^{3\times3}\)
represents the camera intrinsic matrix,
while \(\mathbf{R}_f^k \in SO(3)\) and
\(\mathbf{t}_f^k \in \mathbb{R}^3\)
describe the rigid transformation from the world coordinate
system to the camera coordinate system.

Because the supporting plane and surrounding environment may
also be reconstructed during the MVS process,
a semantic-guided filtering step is performed to retain only
points belonging to the artifact.
The basic idea is to reproject each reconstructed 3D point
into all images and evaluate whether the projected pixel
falls inside the artifact mask.

For a 3D point \(\mathbf{X} \in \mathcal{P}_f\),
its coordinates in the camera coordinate system of view \(k\)
are first obtained through the rigid transformation
\begin{equation}
\mathbf{X}^{c}_{f,k}
=
\mathbf{R}_f^k \mathbf{X} + \mathbf{t}_f^k,
\end{equation}
where \(\mathbf{X}^{c}_{f,k} = [x^{c}_{f,k}, y^{c}_{f,k}, z^{c}_{f,k}]^\top\). Using the standard perspective projection model, the normalized image coordinates are computed as
\begin{equation}
\mathbf{x}^{n}_{f,k}
=
\begin{bmatrix}
x^{c}_{f,k}/z^{c}_{f,k} \\
y^{c}_{f,k}/z^{c}_{f,k}
\end{bmatrix}.
\end{equation}
With the camera intrinsic matrix
\begin{equation}
\mathbf{K}^{k}_{f}
=
\begin{bmatrix}
f_x & 0 & c_x \\
0 & f_y & c_y \\
0 & 0 & 1
\end{bmatrix},
\end{equation}
the projected pixel coordinate is obtained by
\begin{equation}
\mathbf{u}_{f,k}
=
\begin{bmatrix}
u_{f,k} \\
v_{f,k}
\end{bmatrix}
=
\begin{bmatrix}
f_x x^{c}_{f,k}/z^{c}_{f,k} + c_x \\
f_y y^{c}_{f,k}/z^{c}_{f,k} + c_y
\end{bmatrix}.
\end{equation}

Let \(\Omega_f^k \subset \mathbb{R}^2\)
denote the image domain of view \(k\).
A projection is considered geometrically valid only if the
point lies in front of the camera and its projection falls
inside the image domain.
We therefore define the visibility indicator function
\begin{equation}
\delta_{f,k}(\mathbf{X})
=
\mathbf{1}(z^{c}_{f,k} > \epsilon_z)
\cdot
\mathbf{1}(\mathbf{u}_{f,k} \in \Omega_f^k),
\end{equation}
where \(\epsilon_z > 0\) is a small positive constant used to
exclude points located behind the camera,
and \(\mathbf{1}(\cdot)\) denotes the indicator function.

For each valid projection, the binary artifact mask
\(M_f^k : \Omega_f^k \rightarrow \{0,1\}\)
is queried at the projected pixel location
\begin{equation}
m_{f,k}(\mathbf{X}) = M_f^k(\mathbf{u}_{f,k}).
\end{equation}

Using the visibility indicator, the number of valid
observations of point \(\mathbf{X}\) is defined as
\begin{equation}
N_{\mathrm{vis}}^{(f)}(\mathbf{X})
=
\sum_{k=1}^{N_f}
\delta_{f,k}(\mathbf{X}),
\end{equation}
while the number of observations that support the hypothesis
that the point belongs to the artifact is
\begin{equation}
N_{\mathrm{art}}^{(f)}(\mathbf{X})
=
\sum_{k=1}^{N_f}
\delta_{f,k}(\mathbf{X})
m_{f,k}(\mathbf{X}).
\end{equation}

Based on these statistics, we define a semantic consistency
score
\begin{equation}
S_f(\mathbf{X})
=
\frac{
N_{\mathrm{art}}^{(f)}(\mathbf{X})
}{
N_{\mathrm{vis}}^{(f)}(\mathbf{X}) + \varepsilon
},
\end{equation}
where \(\varepsilon>0\) is a small constant introduced
to avoid division by zero.

Intuitively, \(S_f(\mathbf{X})\) measures the proportion
of views in which the projected point is classified as
belonging to the artifact.
Points that receive consistent support across multiple
views are therefore more likely to correspond to the
artifact surface.

Finally, points whose semantic consistency score exceeds
a predefined threshold \(\tau\) are retained
\begin{equation}
\widetilde{\mathcal{P}}_{f}^{\mathrm{sem}}
=
\{\mathbf{X}\in\mathcal{P}_f
\mid
S_f(\mathbf{X}) \ge \tau
\}.
\end{equation}

This multi-view semantic filtering procedure effectively
removes points reconstructed from the supporting plane
and surrounding background while preserving geometrically
consistent artifact points.

The same procedure is applied independently to the back-side
image sequence.
Let
\begin{equation}
\mathcal{I}_b=\{I_b^k\}_{k=1}^{N_b},\quad
\mathcal{M}_b=\{M_b^k\}_{k=1}^{N_b},\quad
\mathcal{P}_b=\{\mathbf{X}_i^{(b)}\in\mathbb{R}^3\}_{i=1}^{M_b},
\end{equation}
denote the back-side images, artifact masks, and reconstructed
dense point cloud, respectively, with corresponding camera
parameters
\begin{equation}
\{\mathbf{K}_b^k,\mathbf{R}_b^k,\mathbf{t}_b^k\}_{k=1}^{N_b}.
\end{equation}

Following the same projection and visibility formulation
defined for the front side, each point
\(\mathbf{X}\in\mathcal{P}_b\) is reprojected into every image
to determine whether it is visible and whether the projected
pixel belongs to the artifact mask.

The number of valid observations and artifact-supporting
observations are defined as
\begin{align}
N_{\mathrm{vis}}^{(b)}(\mathbf{X})
&=
\sum_{k=1}^{N_b}\delta_{b,k}(\mathbf{X}), \\
N_{\mathrm{art}}^{(b)}(\mathbf{X})
&=
\sum_{k=1}^{N_b}\delta_{b,k}(\mathbf{X})\,m_{b,k}(\mathbf{X}),
\end{align}
where \(\delta_{b,k}(\mathbf{X})\) denotes the visibility
indicator and \(m_{b,k}(\mathbf{X})\) is the artifact mask value
sampled at the projected pixel.

The semantic consistency score for the back-side point cloud
is therefore
\begin{equation}
S_b(\mathbf{X}) =
\frac{
N_{\mathrm{art}}^{(b)}(\mathbf{X})
}{
N_{\mathrm{vis}}^{(b)}(\mathbf{X})+\varepsilon
}.
\end{equation}

Finally, points whose score exceeds a threshold \(\tau\)
are retained:
\begin{equation}
\tilde{\mathcal{P}}_b^{\,\mathrm{sem}}=
\left\{
\mathbf{X}\in\mathcal{P}_b
\mid
S_b(\mathbf{X})\ge\tau
\right\}.
\end{equation}

The threshold \(\tau\) controls the strictness of mask-guided point cloud purification. In all experiments, we set \(\tau=0.9\) to retain points with highly consistent foreground support across valid projections. This setting is used to suppress residual background and boundary artifacts, and its influence is evaluated in Section~4.3.1.

Although semantic masking effectively suppresses a large portion of the supporting plane and surrounding background, the reconstructed point clouds may still contain geometric artifacts and sparse outliers introduced by the SfM--MVS pipeline. In particular, weakly textured regions, grazing-angle observations, and imperfect depth fusion near the artifact boundary can generate isolated floating points or small background fragments that remain attached to the artifact surface. Therefore, beyond the semantic consistency filtering in \(\tilde{\mathcal{P}}_s^{\,\mathrm{sem}}\) (\(s\in\{f,b\}\)), we further enforce geometric cleanliness via a structured 3D purification procedure that removes statistical outliers and retains only the dominant connected structure corresponding to the artifact.

We first apply Statistical Outlier Removal (SOR) to \(\tilde{\mathcal{P}}_s^{\,\mathrm{sem}}\). For a point \(\mathbf{X}\in\tilde{\mathcal{P}}_s^{\,\mathrm{sem}}\), let \(\mathcal{N}_K^{(s)}(\mathbf{X})\) denote its \(K\)-nearest neighbors, and define the local mean distance
\begin{equation}
\bar{d}_s(\mathbf{X})=
\frac{1}{K}\sum_{\mathbf{Y}\in \mathcal{N}_K^{(s)}(\mathbf{X})}\|\mathbf{X}-\mathbf{Y}\|_2.
\label{eq:sor_mean}
\end{equation}
Let \(\mu_s\) and \(\sigma_s\) be the mean and standard deviation of \(\{\bar{d}_s(\mathbf{X})\}\) over \(\tilde{\mathcal{P}}_s^{\,\mathrm{sem}}\). Points whose neighborhood statistics deviate significantly from the global distribution are regarded as outliers and removed by
\begin{equation}
\tilde{\mathcal{P}}_s^{\,\mathrm{sor}}=
\left\{\mathbf{X}\in\tilde{\mathcal{P}}_s^{\,\mathrm{sem}}
\;\middle|\;
\bar{d}_s(\mathbf{X})\le \mu_s+\alpha\sigma_s
\right\}, \qquad s\in\{f,b\},
\label{eq:sor_keep}
\end{equation}
where \(\alpha>0\) controls the trimming strength.

After statistical outlier removal, the point cloud may still contain
residual background structures that remain attached to the artifact
surface. These structures typically appear as small connected fragments
or thin surface patches originating from partial reconstruction of the
supporting plane near the artifact boundary. Because such remnants may
form locally coherent structures rather than isolated points, they are
not reliably eliminated by purely local statistical filtering.

To address this issue, we further enforce structural consistency by
retaining only the dominant connected component corresponding to the
artifact geometry. Since clustering directly on the full-resolution
point cloud can be computationally expensive, we first construct a
coarse geometric representation through voxel downsampling. Let
\(\mathcal{D}(\cdot)\) denote the voxel downsampling operator and define
\begin{equation}
\mathcal{Q}_s = \mathcal{D}(\tilde{\mathcal{P}}_s^{\,\mathrm{sor}}).
\end{equation}

We then perform density-based spatial clustering on the downsampled
point set \(\mathcal{Q}_s\) using DBSCAN, which partitions the points
into spatially connected clusters
\(\{\mathcal{C}_{s,m}\}_{m=1}^{L_s}\) according to a neighborhood radius
and a minimum density criterion.

After semantic filtering and statistical denoising, the artifact
typically forms the largest and most coherent connected structure in
the scene. We therefore identify the dominant cluster by selecting the
component with the maximum cardinality
\begin{equation}
\mathcal{C}_{s,*}
=
\arg\max_{m}\,|\mathcal{C}_{s,m}|,
\qquad s\in\{f,b\}.
\label{eq:dbscan_main}
\end{equation}

Finally, the selected structural component is propagated back to the
original-resolution point cloud. Specifically, we retain points in
\(\tilde{\mathcal{P}}_s^{\,\mathrm{sor}}\) whose distance to the dominant
cluster is smaller than a spatial tolerance \(\eta>0\):
\begin{equation}
\tilde{\mathcal{P}}_s=
\left\{
\mathbf{X}\in\tilde{\mathcal{P}}_s^{\,\mathrm{sor}}
\;\middle|\;
\min_{\mathbf{Y}\in \mathcal{C}_{s,*}}
\|\mathbf{X}-\mathbf{Y}\|_2 < \eta
\right\}, \qquad s\in\{f,b\}.
\label{eq:final_pure}
\end{equation}

The resulting \(\tilde{\mathcal{P}}_f\) and \(\tilde{\mathcal{P}}_b\) are thus purified artifact-only point clouds for the front and back sessions, respectively. Since the above operations are performed independently for each side \(s\in\{f,b\}\), the purification stage does not assume any cross-side correspondences or overlap; instead, it ensures that the subsequent registration is initialized from two structurally consistent and background-suppressed surfaces, which is particularly important under the near-zero overlap condition induced by double-sided acquisition.

\subsection{PCA-Based Canonical Normalization and Thickness Estimation}

After semantic-guided 2D--3D purification, we obtain two artifact-only point clouds
\[
\tilde{\mathcal{P}}_f \subset \mathbb{R}^3,
\qquad
\tilde{\mathcal{P}}_b \subset \mathbb{R}^3,
\]
corresponding to the front and back acquisition sessions. Because the two sides are reconstructed independently, the resulting point sets are defined in different session-specific coordinate systems. Consequently, the relative pose between them is unknown, and small scale inconsistencies may also arise during reconstruction. To stabilize later processing, both point clouds are mapped into canonical coordinate systems defined by their intrinsic geometric structure. This normalization step reduces the ambiguity induced by arbitrary reconstruction axes and provides a consistent reference frame for estimating the physical thickness of the artifact.

For each side \(s \in \{f,b\}\), the purified point cloud is denoted as
\(\tilde{\mathcal{P}}_s=\{\mathbf{X}^{(s)}_i\}_{i=1}^{N_s}\), where \(\mathbf{X}^{(s)}_i\in\mathbb{R}^3\).
We apply PCA to the centered point cloud by computing

\begin{equation}
\mathbf{c}_s=\frac{1}{N_s}\sum_{i=1}^{N_s}\mathbf{X}^{(s)}_i,\quad
\mathbf{\Sigma}_s=
\frac{1}{N_s}\sum_{i=1}^{N_s}
(\mathbf{X}^{(s)}_i-\mathbf{c}_s)(\mathbf{X}^{(s)}_i-\mathbf{c}_s)^{\top}.
\end{equation}

The covariance matrix is decomposed as

\begin{equation}
\mathbf{\Sigma}_s=\mathbf{E}_s\mathbf{\Lambda}_s\mathbf{E}^{\top}_s,\quad
\mathbf{E}_s=[\mathbf{e}_{s,1},\mathbf{e}_{s,2},\mathbf{e}_{s,3}],\quad
\lambda_{s,1}\geq\lambda_{s,2}\geq\lambda_{s,3}\geq0 .
\end{equation}

Here, \(\mathbf{e}_{s,1}\) and \(\mathbf{e}_{s,2}\) represent the dominant surface directions, while \(\mathbf{e}_{s,3}\) corresponds to the least-variance direction and is used as the thickness axis.

For thin sheet-like artifacts, this ordering has a clear geometric interpretation. The eigenvectors associated with the two largest eigenvalues span the dominant surface plane of the artifact, while the eigenvector corresponding to the smallest eigenvalue defines the least-variance direction and therefore approximates the estimated physical thickness axis. In practice, eigen-decomposition may introduce a reflection due to sign ambiguity. Following the implementation, we enforce a right-handed canonical basis by requiring the rotation matrix to satisfy \(\det(\mathbf{E}_s)=1\). If \(\det(\mathbf{E}_s)<0\), the sign of one axis (typically \(\mathbf{e}_{s,3}\)) is flipped so that the resulting canonical frame is a proper rotation and preserves geometric orientation consistency.

Using this observation, a canonical coordinate transformation is defined as
\begin{equation}
\mathbf{X}^{\mathrm{pca}}_{s}(\mathbf{X})
=
\mathbf{E}_s^\top
(\mathbf{X}-\mathbf{c}_s),
\qquad
\mathbf{X}\in\tilde{\mathcal{P}}_s .
\label{eq:pca_transform}
\end{equation}

Applying this transformation to all points produces the PCA-normalized point cloud
\begin{equation}
\tilde{\mathcal{P}}_s^{\mathrm{pca}}
=
\left\{
\mathbf{X}^{\mathrm{pca}}_{s}(\mathbf{X}_i^{(s)})
\right\}_{i=1}^{N_s}.
\label{eq:pca_cloud}
\end{equation}

The transformed cloud is centered at the origin and registered with the principal axes of the object geometry. This canonicalization has two practical consequences. First, it makes the dominant surface directions comparable across sessions despite arbitrary reconstruction axes. Second, the third canonical coordinate is consistent with the estimated thickness axis, enabling thickness statistics to be computed by a one-dimensional analysis along this axis.

Since independent reconstructions may exhibit scale drift, we additionally estimate a robust isotropic scale factor between the two PCA-normalized clouds using radial statistics in canonical space. Let
\[
\rho_{s,i}=\left\|
\mathbf{X}^{\mathrm{pca}}_{s}(\mathbf{X}_i^{(s)})
\right\|_2,
\qquad
\mathcal{R}_s=\{\rho_{s,i}\}_{i=1}^{N_s}.
\]
We define the scale factor that maps the back-side canonical cloud to the front-side canonical scale as
\begin{equation}
\alpha
=
\frac{\mathrm{perc}_{90}(\mathcal{R}_f)}{\mathrm{perc}_{90}(\mathcal{R}_b)} .
\label{eq:scale_ratio}
\end{equation}
Using the 90th percentile rather than the maximum avoids sensitivity to isolated outliers or residual floating points, and matches the robust-radius strategy used in the implementation.

Let
\begin{equation}
\mathbf{X}^{\mathrm{pca}}_{s}(\mathbf{X}_i^{(s)})
=
\begin{bmatrix}
x_{s,i}^{\mathrm{pca}} &
y_{s,i}^{\mathrm{pca}} &
z_{s,i}^{\mathrm{pca}}
\end{bmatrix}^\top ,
\label{eq:pca_coord}
\end{equation}
and denote the set of thickness-axis coordinates as
\begin{equation}
\mathcal{Z}_s=\{z_{s,i}^{\mathrm{pca}}\}_{i=1}^{N_s}.
\label{eq:z_set}
\end{equation}

Directly using the minimum and maximum values of \(\mathcal{Z}_s\) is sensitive to reconstruction noise and extreme artifacts near grazing angles. To obtain a robust estimate, we compute percentile-based bounds
\begin{equation}
z_s^{\mathrm{low}}=\mathrm{perc}_2(\mathcal{Z}_s),
\qquad
z_s^{\mathrm{high}}=\mathrm{perc}_{98}(\mathcal{Z}_s).
\label{eq:percentile_bounds}
\end{equation}

The 2nd--98th percentile interval is used as a robust empirical setting for thickness-axis extent estimation. This interval excludes sparse extreme points caused by reconstruction noise, boundary artifacts, or grazing-angle observations, while preserving the main distribution of the front and back surfaces along the thickness axis. The influence of this parameter is further evaluated in Section~4.3.2.

The thickness extent of each side is then defined as
\begin{equation}
\Delta_s
=
z_s^{\mathrm{high}}
-
z_s^{\mathrm{low}},
\qquad
s\in\{f,b\}.
\label{eq:side_thickness}
\end{equation}

To account for scale drift, we aggregate the two side-specific extents after bringing them to a common scale. Specifically, the back-side thickness extent is scaled by \(\alpha\), yielding a global thickness estimate
\begin{equation}
h=
\frac{\Delta_f+\alpha\,\Delta_b}{2}.
\label{eq:global_thickness}
\end{equation}
This definition is consistent with the implementation logic that first normalizes scale using a robust radius statistic and then evaluates thickness using canonical \(z\)-axis variation. The resulting estimate \(h\) provides a physically meaningful thickness proxy that is stable under mild reconstruction noise and inter-session scaling differences.

Finally, the thickness estimate is converted into an explicit separation magnitude along the canonical thickness axis:
\begin{equation}
z_{\mathrm{off}}=\frac{1}{2}h .
\label{eq:z_offset}
\end{equation}
The value \(z_{\mathrm{off}}\) is retained as a thickness-derived geometric prior for downstream use. It encodes a physically plausible half-thickness separation in the canonical coordinate system and is derived solely from intrinsic statistics of the purified point clouds, without assuming any cross-side correspondences.

\subsection{Geometry-Constrained Bidirectional Registration}

After PCA-based canonical normalization, the purified front and back point clouds are represented in their respective PCA coordinate systems:

\begin{equation}
\tilde{\mathcal{P}}_f^{\mathrm{pca}}
=
\left\{
\mathbf{X}_{f,i}^{\mathrm{pca}}
\right\}_{i=1}^{N_f},
\qquad
\tilde{\mathcal{P}}_b^{\mathrm{pca}}
=
\left\{
\mathbf{X}_{b,j}^{\mathrm{pca}}
\right\}_{j=1}^{N_b}.
\end{equation}

In this canonical representation, both point clouds are centered at the origin and registered with their principal axes. 
For thin sheet-like objects, the third PCA axis corresponds to the direction of minimum geometric variance and therefore approximates the physical thickness direction of the artifact. 
This canonicalization significantly reduces the pose ambiguity between the independently reconstructed front and back surfaces.

\begin{figure}[t]
    \centering
    \includegraphics[width=\linewidth]{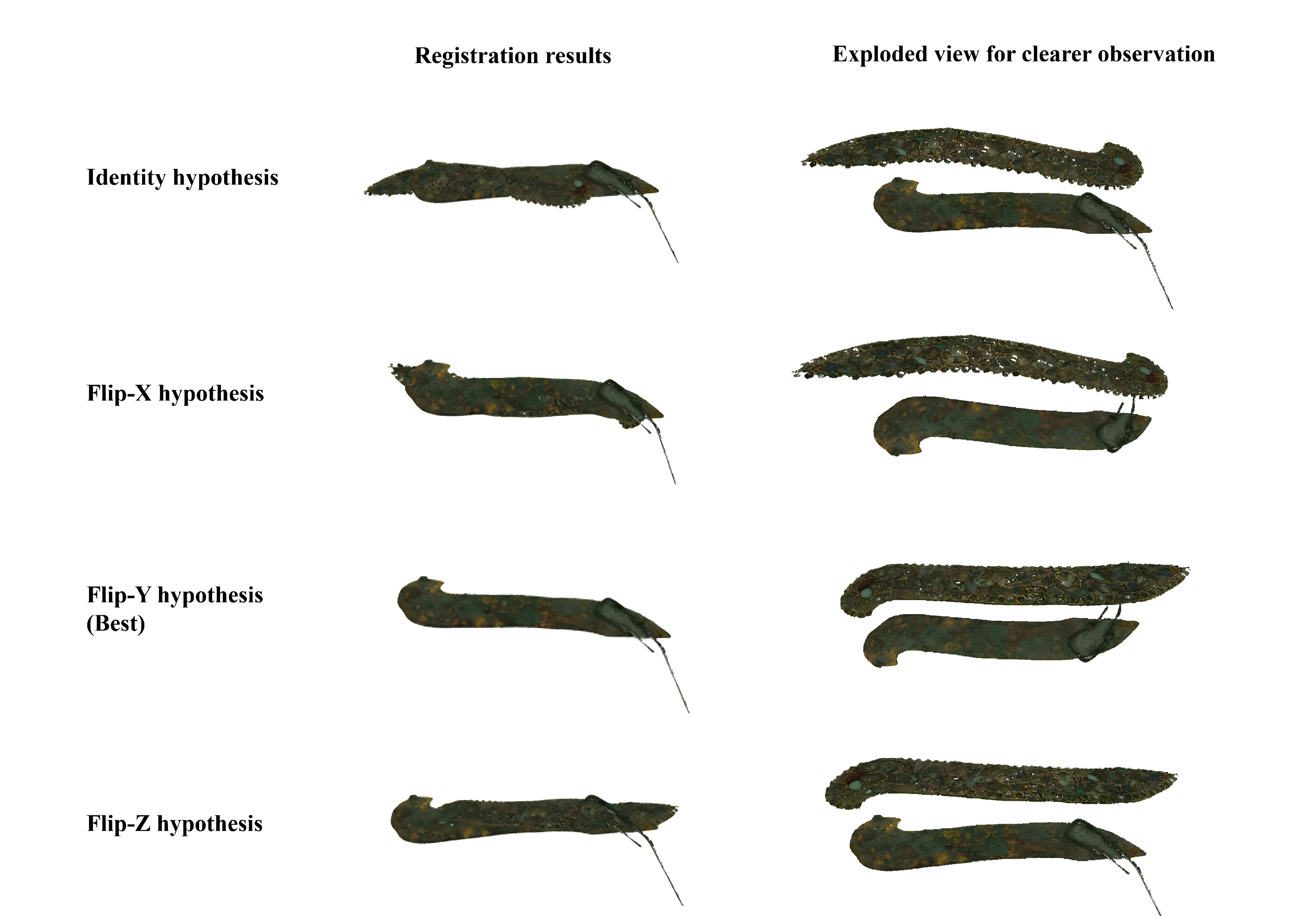}
    \caption{Visualization of point cloud registration under different hypotheses. 
    The first column shows the direct registration results of the two point clouds. 
    The second column presents an exploded view where the two point clouds are spatially separated with a large offset to better observe their relative rotations. 
    The best alignment hypothesis is highlighted.}
    \label{fig:registration_hypotheses}
    \Description{Four registration hypotheses for the two surfaces of an elongated decorated artifact are compared in rows: identity, flip about the X axis, flip about the Y axis, and flip about the Z axis. The left column shows the merged registration result, while the right column separates the two surfaces vertically for inspection. The flip-Y hypothesis produces the most consistent orientation and boundary correspondence and is identified as the best result.}
\end{figure}

However, eigenvectors obtained from PCA are defined only up to sign. 
As a result, the PCA coordinate frames of the two surfaces may differ by axis flips, which may cause the back surface to appear mirrored relative to the front surface along one or more principal axes. 
To resolve this ambiguity, we evaluate a finite set of global orientation hypotheses. As shown in Fig.~\ref{fig:registration_hypotheses}, different axis-flip hypotheses lead to different point cloud registration results.

For thin sheet-like artifacts captured through double-sided acquisition, the physically plausible configurations can be restricted to $180^\circ$ rotations around the canonical axes. 
This yields the following hypothesis set:

\begin{equation}
\mathcal{R}
=
\left\{
\mathbf{I},
\mathbf{R}_x(\pi),
\mathbf{R}_y(\pi),
\mathbf{R}_z(\pi)
\right\}.
\end{equation}

\begin{equation}
\mathbf{R}_x(\pi)=
\begin{bmatrix}
1 & 0 & 0 \\
0 & -1 & 0 \\
0 & 0 & -1
\end{bmatrix},
\qquad
\mathbf{R}_y(\pi)=
\begin{bmatrix}
-1 & 0 & 0 \\
0 & 1 & 0 \\
0 & 0 & -1
\end{bmatrix},
\qquad
\mathbf{R}_z(\pi)=
\begin{bmatrix}
-1 & 0 & 0 \\
0 & -1 & 0 \\
0 & 0 & 1
\end{bmatrix}.
\end{equation}

Each $\mathbf{R}_k \in \mathcal{R}$ defines a candidate orientation of the back surface relative to the front surface in PCA space.

Before ICP refinement, each rotation hypothesis is initialized with the thickness-derived offset \(z_{\mathrm{off}}\) defined in Eq.~\eqref{eq:z_offset}, which separates the two surfaces along the canonical thickness axis:

\begin{equation}
\mathbf{T}^{(0)}_k =
\begin{bmatrix}
\mathbf{R}_k & \mathbf{t}^{(0)}_k \\
\mathbf{0}^{\top} & 1
\end{bmatrix},
\quad
\mathbf{t}^{(0)}_k =
\begin{bmatrix}
0 \\
0 \\
z_{\mathrm{off}}
\end{bmatrix},
\quad
z_{\mathrm{off}}=\frac{1}{2}h .
\end{equation}

The initialized back-side point cloud under hypothesis \(\mathbf{R}_k\) is then given by

\begin{equation}
\hat{\mathcal{P}}^{(0)}_{b,k}
=
\left\{
\mathbf{R}_k \mathbf{X}^{\mathrm{pca}}_{b,j}
+
\mathbf{t}^{(0)}_k
\right\}_{j=1}^{N_b}.
\end{equation}

Starting from the initialized configuration, each hypothesis is refined by point-to-plane ICP, with the front surface used as the fixed reference and the transformed back surface as the source. The optimization minimizes

\begin{equation}
E_k(\mathbf{R},\mathbf{t})
=
\sum_{j=1}^{N_b}
\left(
\mathbf{n}_j^\top
\left(
\mathbf{R}\mathbf{x}_j+\mathbf{t}-\mathbf{y}_j
\right)
\right)^2,
\quad
\mathbf{x}_j=\mathbf{R}_k\mathbf{X}^{\mathrm{pca}}_{b,j}+\mathbf{t}^{(0)}_k .
\end{equation}

After ICP refinement, the resulting transformation for hypothesis \(k\) is written as

\begin{equation}
\mathbf{T}_{\mathrm{ICP},k}
=
\begin{bmatrix}
\mathbf{R}_{\mathrm{ICP},k} & \mathbf{t}_{\mathrm{ICP},k} \\
\mathbf{0}^{\top} & 1
\end{bmatrix},
\quad
\mathbf{T}_k=\mathbf{T}_{\mathrm{ICP},k}\mathbf{T}^{(0)}_k .
\end{equation}

After ICP refinement, the quality of each hypothesis is evaluated using a thickness-aware fitness criterion. 
Specifically, correspondences whose residual distance is smaller than a thickness-scaled threshold are considered inliers:

\begin{equation}
\mathcal{I}_k
=
\left\{
j
\;\middle|\;
\left\|
\mathbf{T}_k
\mathbf{X}_{b,j}^{\mathrm{pca}}
-
\mathbf{y}_j
\right\|_2
<
d_{\max}
\right\},
\qquad
\mathrm{Fitness}_k
=
\frac{|\mathcal{I}_k|}{N_b}.
\end{equation}

The threshold is defined as

\begin{equation}
d_{\max} = \gamma h.
\end{equation}

In our implementation, \(\gamma\) is set to 2.0 unless otherwise specified. This value is selected as an empirical balance between tolerance and discriminability in the thickness-scaled fitness evaluation. A smaller \(\gamma\) makes the correspondence threshold overly strict and may reject valid correspondences affected by reconstruction noise, boundary artifacts, or residual ICP errors. A larger \(\gamma\) makes the threshold overly permissive and may cause fitness saturation among different rotation hypotheses, thereby weakening the ability to distinguish the correct orientation. The sensitivity of \(\gamma\) is analyzed in Section~4.3.3.

Among all hypotheses, the transformation that produces the highest fitness score is selected as the final solution:

\begin{equation}
k^* = \arg\max_{k} \mathrm{Fitness}_k ,
\qquad
\mathbf{T}_{b \to f}^{\mathrm{pca}} = \mathbf{T}_{k^*}.
\end{equation}

Finally, the optimal transformation obtained in PCA space is mapped back to the original world coordinate systems of the two reconstruction sessions:

\begin{equation}
\mathbf{X}_{s}^{\mathrm{pca}} = \mathbf{E}_s^\top (\mathbf{X}-\mathbf{c}_s),
\qquad
\mathbf{X}
=
\mathbf{E}_s \mathbf{X}_{s}^{\mathrm{pca}} + \mathbf{c}_s.
\end{equation}

\begin{equation}
\mathbf{T}_{b \to f}^{\mathrm{world}}
=
\left[\begin{smallmatrix}
\mathbf{E}_f & \mathbf{c}_f \\
\mathbf{0}^{\top} & 1
\end{smallmatrix}\right]
\mathbf{T}_{b \to f}^{\mathrm{pca}}
\left[\begin{smallmatrix}
\mathbf{E}_b^\top & -\mathbf{E}_b^\top\mathbf{c}_b \\
\mathbf{0}^{\top} & 1
\end{smallmatrix}\right].
\end{equation}

Applying $\mathbf{T}_{b \to f}^{\mathrm{world}}$ to $\tilde{\mathcal{P}}_b$ registers the back point cloud with the front surface in the same coordinate system, producing the final stitched artifact model.

\begin{figure}[t]
\centering
\scalebox{1}[1.01]{%
\includegraphics[width=.95\linewidth]{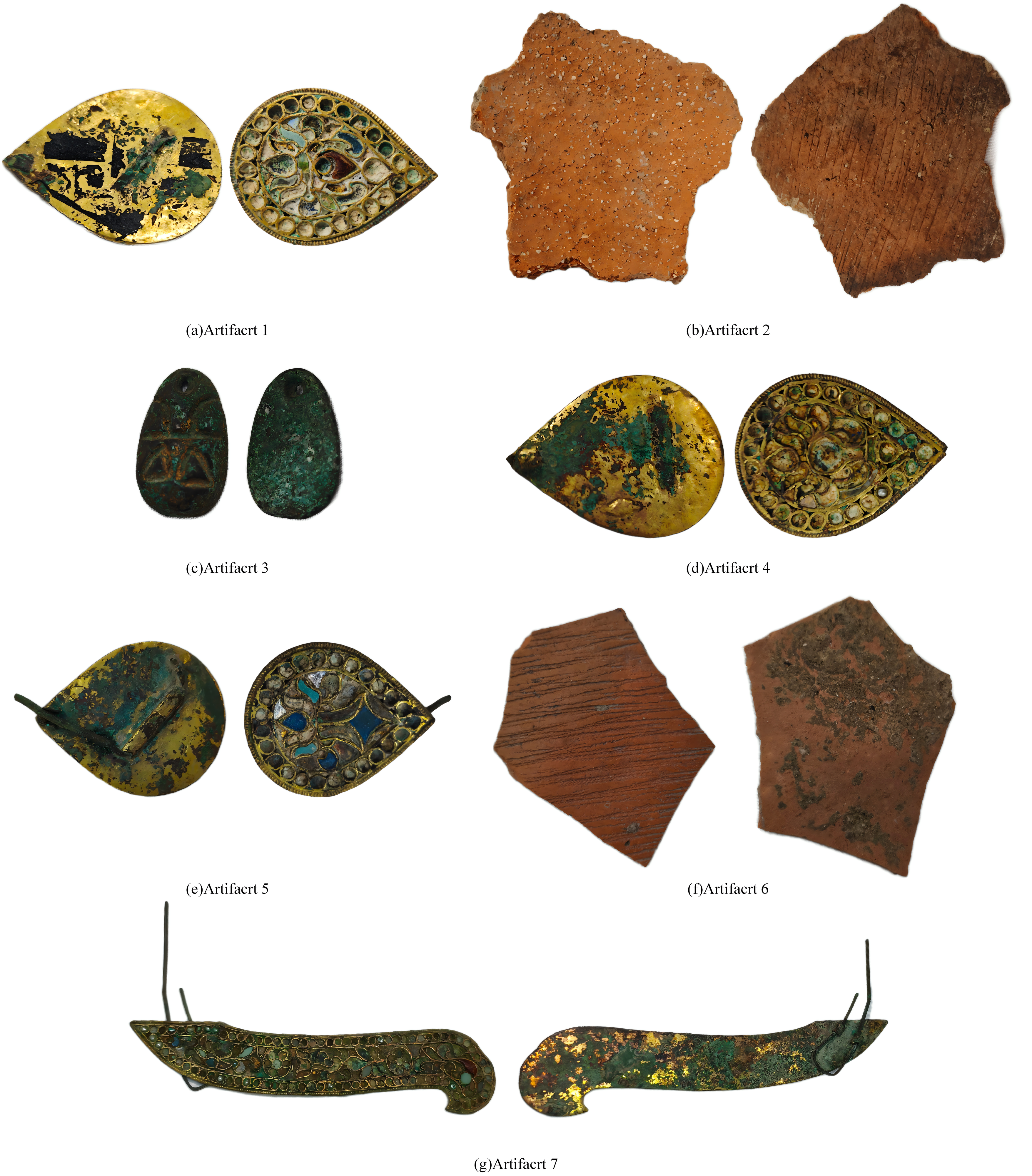}
}
\caption{Front and back views of all artifacts involved in the experiments.}
\label{fig:objects}
\Description{Front and back photographs of seven thin cultural-heritage artifacts arranged as labeled pairs. The collection includes small teardrop-shaped decorated metal pieces, irregular reddish-brown fragments, a dark oval pendant-like object, polygonal textured fragments, and a long narrow decorated metal object. The paired views reveal substantial differences in texture, ornamentation, boundary shape, surface condition, and thickness among the test objects.}
\end{figure}

\section{Experimental Results}

\subsection{Experimental Setup}

The artifact samples were digitized through independent front-side and back-side acquisition sessions. During each session, the artifact was placed on a flat supporting surface and captured from multiple viewpoints, following the non-contact acquisition requirement for fragile heritage materials.

Images were captured using a Redmi K70 Pro smartphone camera. For each side, the camera was moved around the artifact to acquire a multi-view image sequence with sufficient overlap for stable SfM--MVS reconstruction. The front-side and back-side image sequences were processed independently, producing two dense point clouds corresponding to the two surfaces of the artifact.

The reconstructed point clouds were then used as input for the registration experiments. Semantic masks for background removal were generated using SAM2, and the mask inference was performed on a workstation equipped with an NVIDIA RTX 5090D GPU. The same hardware and computational environment were used for subsequent point cloud purification, registration, and baseline comparison to ensure consistent experimental conditions.

The proposed method and all baseline approaches were evaluated on the same set of purified front--back point cloud pairs obtained from the above acquisition and reconstruction workflow.

The artifact samples used in the experiments are shown in Fig.~\ref{fig:objects}. For each object, both front-side and back-side views are presented to illustrate the physical samples considered in the evaluation. These samples represent typical flat and fragile cultural heritage artifacts that require independent double-sided acquisition during non-contact digitization.

\subsection{Comparison with Existing Registration Methods}

To assess the effectiveness of the proposed registration method, we compare it with six representative rigid point cloud registration approaches, including Coherent Point Drift (CPD)~\cite{myronenko2010cpd} followed by Iterative Closest Point (ICP), Fast Global Registration (FGR)~\cite{zhou2016fgr} with ICP refinement, Fast Point Feature Histogram (FPFH)-based registration with ICP refinement, Generalized ICP (GICP), Predator, and GeoTransformer. These baselines cover probabilistic registration, feature-based global alignment, iterative geometric optimization, and learning-based correspondence estimation.

It should be noted that these methods were originally developed for general point cloud registration scenarios in which reliable correspondences or partially overlapping regions are usually available. In contrast, the task considered in this paper focuses on front--back registration of thin, sheet-like heritage artifacts, where the two reconstructed surfaces often share very limited common geometry. Therefore, the comparison is intended to examine the applicability of representative registration methods under this challenging domain-specific setting, rather than to negatively assess their performance in their original target scenarios.

All methods are evaluated using the same purified front--back point cloud pairs described in Section~4.1. Each baseline first estimates a rigid transformation according to its original formulation. To ensure a consistent evaluation protocol, the estimated transformation is further refined using the same point-to-plane ICP procedure. For the learning-based methods, we follow the official implementations and use the released pretrained weights for evaluation.

Since the objective of this work is to obtain a registered front--back model that is consistent with the original image observations, we evaluate the registration results using a rendering-based evaluation procedure. Specifically, each registered point cloud is rendered from selected calibrated viewpoints using the camera poses estimated during reconstruction. The resulting image-space representations, including rendered images and foreground silhouettes, are then compared with the corresponding captured images and object masks derived from the acquisition data.

This rendering-based evaluation procedure is consistent with the practical acquisition setting of thin, sheet-like heritage artifacts. In this study, the available observations mainly consist of calibrated multi-view images acquired under non-contact conditions. Obtaining an additional independent geometric reference model would require extra scanning, fabrication, and calibration procedures. Active scanning or replica-based validation may provide useful complementary evidence, but the accuracy of the resulting reference model can also be affected by surface accessibility, boundary stability, scanning resolution, and fabrication quality. Therefore, this study focuses on evaluating whether the registered and rendered model remains consistent with the original image observations, while more extensive reference-based 3D geometric validation is left for future work.

Based on this evaluation procedure, we employ both image-space metrics and projection-based geometric metrics. PSNR, SSIM, and LPIPS are used to evaluate the visual fidelity between rendered and captured images. IoU and AreaRatio are further used to assess silhouette overlap and projected area consistency. Although these metrics cannot replace direct 3D geometric measurements, they provide practical evidence for contour alignment, projected scale consistency, and image-plane geometric agreement under the available acquisition conditions.

Peak Signal-to-Noise Ratio (PSNR)~\cite{hore2010psnr} is used to measure the pixel-level fidelity between the rendered image $\hat{I}$ and the corresponding reference image $I$. It is defined as
\begin{equation}
PSNR = 10 \log_{10}\left(\frac{MAX_I^2}{MSE}\right),
\end{equation}
where $MAX_I$ denotes the maximum possible pixel intensity, and the mean squared error (MSE) is computed as
\begin{equation}
MSE = \frac{1}{N} \sum_{i=1}^{N} \left(I_i - \hat{I}_i\right)^2 .
\end{equation}

Structural Similarity Index (SSIM)~\cite{wang2004ssim} evaluates the structural consistency between two images by considering luminance, contrast, and structural information. For two image patches $x$ and $y$, SSIM is defined as
\begin{equation}
SSIM(x,y) =
\frac{(2\mu_x \mu_y + C_1)(2\sigma_{xy} + C_2)}
{(\mu_x^2 + \mu_y^2 + C_1)(\sigma_x^2 + \sigma_y^2 + C_2)},
\end{equation}
where $\mu_x$ and $\mu_y$ denote the mean intensities of $x$ and $y$, $\sigma_x^2$ and $\sigma_y^2$ denote the corresponding variances, and $\sigma_{xy}$ denotes their covariance.

Learned Perceptual Image Patch Similarity (LPIPS)~\cite{zhang2018lpips} is adopted to evaluate perceptual similarity in a deep feature space. Let $\phi_l(\cdot)$ denote the feature map extracted from layer $l$. LPIPS is computed as
\begin{equation}
LPIPS(x,y) =
\sum_l \frac{1}{H_l W_l}
\sum_{h,w}
\left\| w_l \odot \left(\phi_l(x)_{h,w} - \phi_l(y)_{h,w}\right) \right\|_2^2 ,
\end{equation}
where $w_l$ denotes the learned channel-wise weights, and $(h,w)$ indexes the spatial locations of the feature map.

\begin{figure}[!htbp]
\centering
\includegraphics[width=1\linewidth]{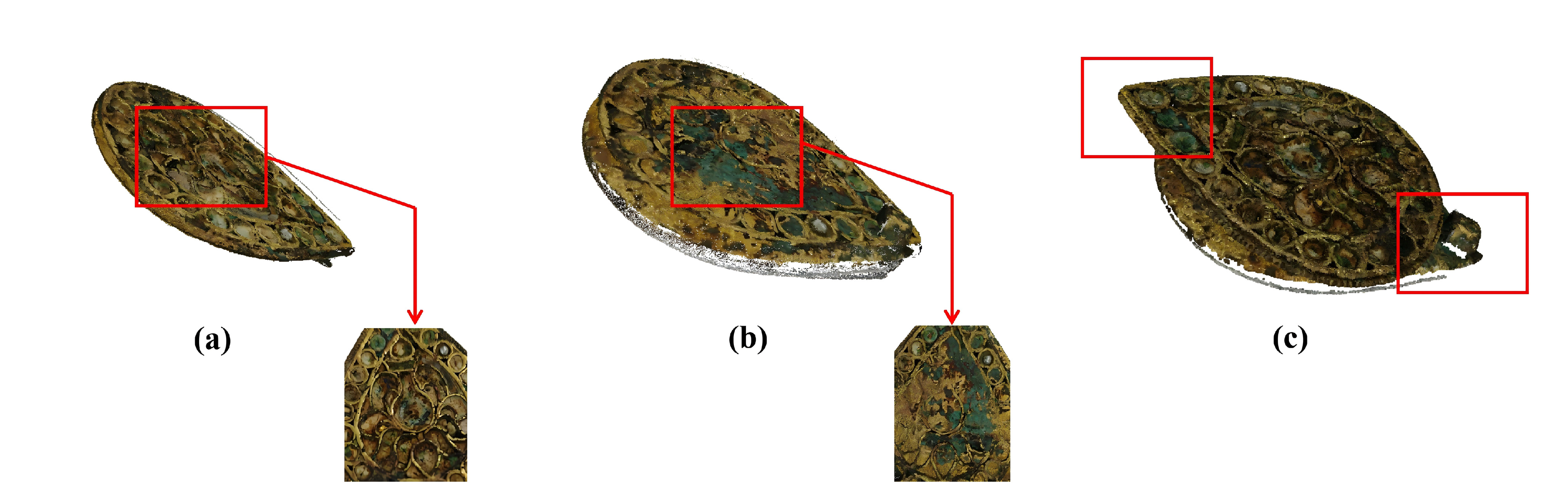}
\caption{Visual comparison of reconstruction results. (a) Result produced by the proposed method. (b) Reconstruction without thickness constraints, resulting in surface interpenetration. (c) Reconstruction without hypothesis flipping, leading to inaccurate stitching.}
\label{fig:comparison_visual}
\Description{Three reconstruction results of the same decorated artifact are shown from left to right. The full proposed method produces a continuous surface with consistent alignment around the highlighted region. Removing the thickness constraint causes the reconstructed front and back surfaces to intersect. Removing rotation-hypothesis evaluation produces an incorrect relative orientation and mismatched local structure. Red boxes and enlarged insets emphasize the differences near the registration boundary.}
\end{figure}

In addition to image-space similarity, silhouette consistency is evaluated by comparing the rendered projection mask with the artifact mask extracted from the real image. Let $S_r$ and $S_g$ denote the foreground silhouette regions of the rendered image and the reference image, respectively. The intersection-over-union is defined as
\begin{equation}
IoU = \frac{|S_r \cap S_g|}{|S_r \cup S_g|}.
\end{equation}

To measure projected area consistency, we further introduce the AreaRatio metric. Let $A_r$ and $A_g$ denote the number of foreground pixels in the rendered mask and the reference mask, respectively. AreaRatio is defined as
\begin{equation}
AreaRatio = \frac{A_r}{A_g}.
\end{equation}
A value closer to 1 indicates that the rendered projection has a projected area more consistent with the observed artifact region.

\begin{table*}[!htbp]
\caption{Comparison across different artifact cases. Within each artifact group, the top three results are emphasized by red-colored text and orange-shaded backgrounds with decreasing intensity, where darker shades denote higher-ranked performance.}
\label{tab:artifact_comparison}
\centering
\tiny
\setlength{\tabcolsep}{4.5pt}
\renewcommand{\arraystretch}{0.85}
\resizebox{\textwidth}{!}{%
\begin{tabular}{@{}llC{1.20cm}C{1.20cm}C{1.20cm}C{1.20cm}C{1.60cm}@{}}
\toprule
\textbf{Object} & \textbf{Baseline methods}
& \textbf{PSNR$\uparrow$}
& \textbf{SSIM$\uparrow$}
& \textbf{LPIPS$\downarrow$}
& \textbf{IoU$\rightarrow 1$}
& \textbf{AreaRatio$\rightarrow 1$} \\
\midrule

\multirow{7}{*}{Artifact1}
& FPFH+ICP       & \rankthree{12.812} & \rankthree{0.477} & 0.620 & 0.292 & 3.134 \\
& CPD+ICP        & 10.668 & 0.358 & 0.611 & 0.303 & \ranktwo{0.607} \\
& FGR+ICP        & 10.585 & 0.298 & 0.545 & \ranktwo{0.497} & \rankthree{1.941} \\
& GICP           & 11.648 & 0.434 & 0.564 & 0.305 & 3.338 \\
& GeoTransformer & 12.694 & 0.436 & \rankthree{0.494} & \rankthree{0.486} & 2.379 \\
& Predator       & \ranktwo{13.137} & \ranktwo{0.483} & \ranktwo{0.485} & 0.459 & 2.166 \\
& Ours           & \rankone{15.535} & \rankone{0.536} & \rankone{0.327} & \rankone{0.600} & \rankone{1.075} \\
\midrule

\multirow{7}{*}{Artifact2}
& FPFH+ICP       & 12.601 & 0.395 & 0.558 & 0.510 & 0.510 \\
& CPD+ICP        & 11.193 & 0.409 & 0.586 & 0.376 & 1.376 \\
& FGR+ICP        & \rankthree{14.630} & \ranktwo{0.456} & \rankthree{0.445} & \rankthree{0.734} & \ranktwo{1.038} \\
& GICP           & 13.362 & \rankthree{0.414} & 0.474 & 0.582 & 1.469 \\
& GeoTransformer & 14.412 & \rankthree{0.414} & 0.449 & \ranktwo{0.738} & \rankthree{0.932} \\
& Predator       & \ranktwo{14.839} & \rankone{0.481} & \ranktwo{0.438} & 0.733 & 1.166 \\
& Ours           & \rankone{15.606} & 0.349 & \rankone{0.367} & \rankone{0.757} & \rankone{0.987} \\
\midrule

\multirow{7}{*}{Artifact3}
& FPFH+ICP       & 13.900 & 0.416 & 0.547 & 0.500 & 2.007 \\
& CPD+ICP        & 15.573 & \ranktwo{0.490} & \rankthree{0.545} & 0.186 & \rankthree{1.623} \\
& FGR+ICP        & 15.128 & 0.234 & 0.677 & 0.383 & 2.018 \\
& GICP           & 13.866 & 0.400 & 0.547 & 0.504 & 2.013 \\
& GeoTransformer & \ranktwo{18.116} & \rankthree{0.474} & \ranktwo{0.496} & \rankthree{0.627} & 1.805 \\
& Predator       & \rankthree{15.940} & 0.390 & 0.549 & \ranktwo{0.651} & \ranktwo{1.614} \\
& Ours           & \rankone{20.909} & \rankone{0.619} & \rankone{0.342} & \rankone{0.689} & \rankone{1.073} \\
\midrule

\multirow{7}{*}{Artifact4}
& FPFH+ICP       & \ranktwo{18.235} & \ranktwo{0.750} & \ranktwo{0.301} & \ranktwo{0.840} & 1.680 \\
& CPD+ICP        & 12.452 & 0.318 & 0.704 & 0.378 & 1.504 \\
& FGR+ICP        & \rankthree{15.560} & 0.397 & \rankthree{0.359} & 0.521 & \rankthree{1.254} \\
& GICP           & 13.144 & 0.283 & 0.538 & \rankone{0.885} & 1.445 \\
& GeoTransformer & 15.119 & \rankthree{0.435} & 0.403 & 0.730 & 1.338 \\
& Predator       & 15.413 & 0.350 & 0.455 & \rankthree{0.773} & \ranktwo{1.175} \\
& Ours           & \rankone{18.489} & \rankone{0.762} & \rankone{0.289} & 0.681 & \rankone{1.114} \\
\midrule

\multirow{7}{*}{Artifact5}
& FPFH+ICP       & 14.668 & 0.354 & 0.496 & 0.471 & 1.820 \\
& CPD+ICP        & \rankthree{15.388} & \ranktwo{0.400} & \rankthree{0.438} & 0.496 & \ranktwo{1.144} \\
& FGR+ICP        & \ranktwo{15.439} & \rankthree{0.362} & \ranktwo{0.431} & \ranktwo{0.601} & \rankthree{1.229} \\
& GICP           & 14.910 & 0.357 & 0.445 & \rankthree{0.559} & 1.713 \\
& GeoTransformer & \rankone{15.875} & \rankone{0.403} & \rankone{0.369} & \rankone{0.709} & 1.392 \\
& Predator       & 14.974 & 0.344 & 0.487 & 0.552 & 1.300 \\
& Ours           & 14.602 & 0.286 & 0.497 & 0.438 & \rankone{0.982} \\
\midrule

\multirow{7}{*}{Artifact6}
& FPFH+ICP       & 15.529 & \rankone{0.536} & \rankthree{0.466} & 0.520 & 1.971 \\
& CPD+ICP        & 15.193 & \rankthree{0.509} & 0.522 & 0.370 & 2.029 \\
& FGR+ICP        & \rankthree{16.327} & 0.489 & 0.532 & \rankone{0.772} & \ranktwo{1.254} \\
& GICP           & 15.473 & \ranktwo{0.526} & \rankthree{0.466} & 0.515 & 1.947 \\
& GeoTransformer & \ranktwo{16.710} & 0.474 & \ranktwo{0.463} & \ranktwo{0.717} & 1.453 \\
& Predator       & 15.937 & 0.434 & 0.546 & 0.548 & \rankthree{1.444} \\
& Ours           & \rankone{18.177} & 0.499 & \rankone{0.396} & \rankthree{0.710} & \rankone{1.013} \\
\midrule

\multirow{7}{*}{Artifact7}
& FPFH+ICP       & \rankthree{17.939} & 0.565 & 0.530 & \rankone{1.000} & \rankthree{1.166} \\
& CPD+ICP        & 17.566 & 0.551 & 0.552 & \rankthree{0.456} & \rankone{1.033} \\
& FGR+ICP        & 17.113 & 0.501 & 0.583 & 0.377 & 1.611 \\
& GICP           & \ranktwo{18.204} & \rankthree{0.586} & \rankone{0.496} & 0.228 & 2.240 \\
& GeoTransformer & 17.305 & 0.514 & 0.523 & 0.171 & 1.992 \\
& Predator       & \ranktwo{18.204} & \ranktwo{0.593} & \rankthree{0.522} & 0.088 & 2.151 \\
& Ours           & \rankone{19.666} & \rankone{0.677} & \ranktwo{0.504} & \ranktwo{0.790} & \ranktwo{0.865} \\

\bottomrule
\end{tabular}%
}
\end{table*}

The quantitative comparison results are summarized in Table~\ref{tab:artifact_comparison}, where all metrics are averaged over the evaluated viewpoints. The results show that registration performance varies with artifact shape, thickness, texture, and projected extent. In addition, IoU and AreaRatio provide complementary evidence: AreaRatio reflects projected area consistency, whereas IoU further measures spatial silhouette overlap. The qualitative results in Fig.~\ref{fig:comparison_visual} further demonstrate the effectiveness of the proposed method under challenging thin-sheet registration conditions.

\subsection{Parameter Sensitivity and Robustness Evaluation}

To address the influence of heuristic parameter choices and evaluate the robustness of the proposed pipeline, we conduct additional sensitivity and robustness analyses. Specifically, we analyze three representative parameters used in different stages of the method: the semantic consistency threshold $\tau$ in mask-guided point cloud purification, the percentile thresholds used for thickness estimation, and the $\gamma$ parameter used in the thickness-scaled fitness evaluation. These parameters respectively affect artifact-only point cloud extraction, thickness-axis extent estimation, and rotation hypothesis evaluation. In addition, we conduct a synthetic noise perturbation experiment to evaluate the stability of the proposed registration method under geometric noise. The objects used in these analyses are selected from the experimental dataset as representative examples. The artifacts in our dataset can be roughly divided into four groups according to their size, thickness, texture richness, and material characteristics. The first group includes Artifacts 1, 4, and 5, which are small thin-sheet relics with relatively rich surface textures. The second group includes Artifacts 2 and 6, which are larger and relatively thicker fragments, but their surface textures are comparatively weak. Artifact 3 represents a small compact relic category, with both limited surface area and small thickness, as well as weak texture information. Artifact 7 has material and surface characteristics similar to Artifacts 1, 4, and 5, but it is larger in scale and contains richer texture details.

\subsubsection{Sensitivity to the Semantic Consistency Threshold}

\begin{figure}[!htbp]
\centering
\includegraphics[width=0.9\textwidth]{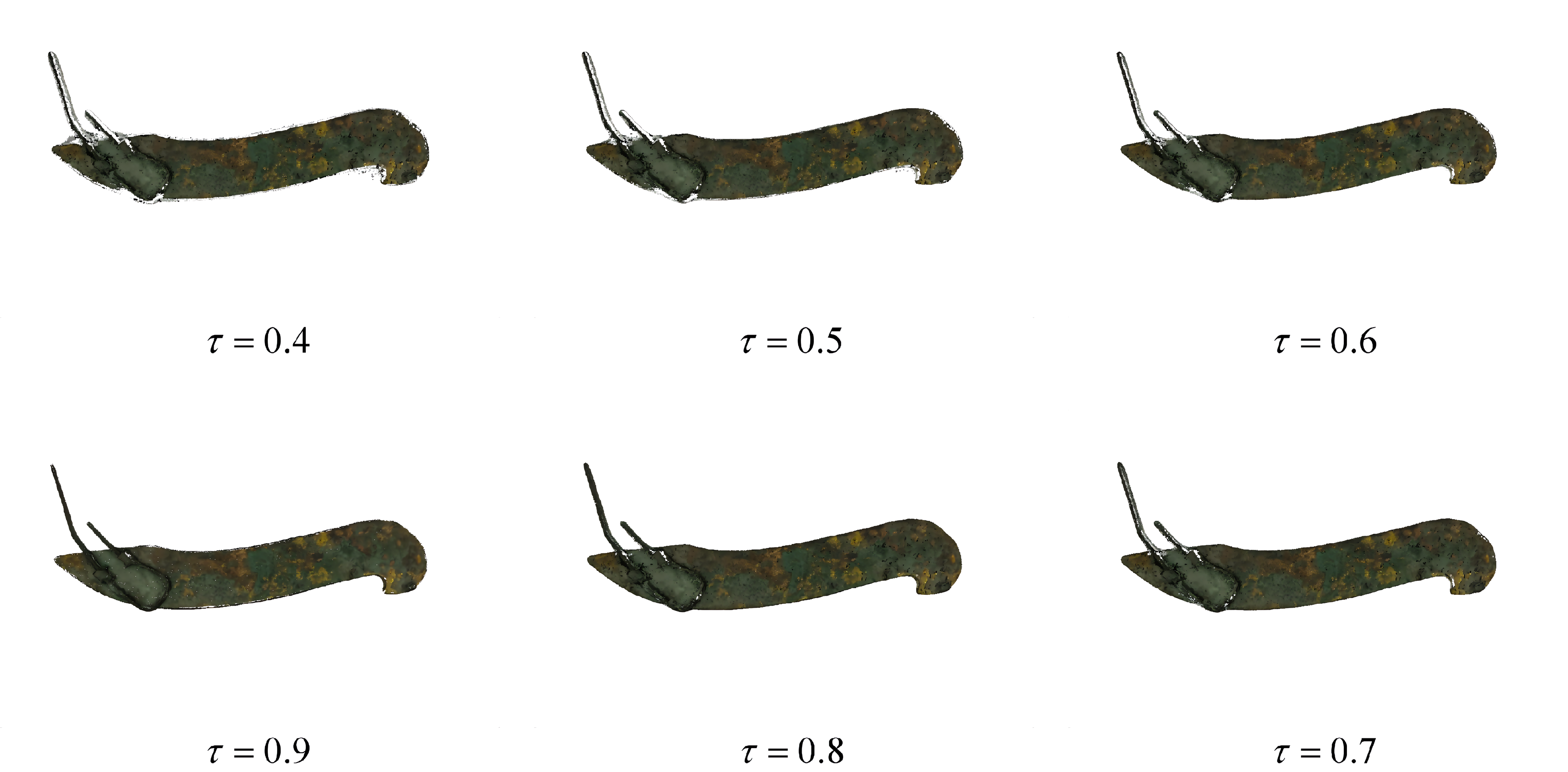}
\caption{Sensitivity of mask-guided point cloud purification to different semantic consistency thresholds $\tau$. As $\tau$ increases from 0.4 to 0.9, the purified point clouds remain visually stable, with only minor changes near the boundary regions.}
\label{fig:tau_sensitivity}
\Description{Six renderings of the same elongated artifact after mask-guided point-cloud purification, corresponding to semantic-consistency thresholds of 0.4, 0.5, 0.6, 0.7, 0.8, and 0.9. The main artifact surface and overall boundary remain nearly unchanged across all thresholds. Increasing the threshold progressively removes a small number of isolated and weakly supported points near the outer boundary while preserving the central structure.}
\end{figure}

The semantic consistency threshold $\tau$ is used in the mask-guided point cloud purification stage. It determines whether a reconstructed 3D point should be retained according to the proportion of valid visible views in which its projection falls inside the artifact mask. A smaller $\tau$ preserves more points but may retain residual background or boundary artifacts, whereas a larger $\tau$ applies a stricter multi-view consistency requirement and removes more uncertain points.
 
To evaluate the influence of this parameter, we test different $\tau$ values from 0.4 to 0.9. The corresponding purification results are shown in Fig.~\ref{fig:tau_sensitivity}. As $\tau$ increases, the retained point cloud becomes slightly cleaner near the object boundary, while the main artifact structure remains stable. This indicates that the mask-guided purification result is not highly sensitive to the specific choice of $\tau$ within this range. In particular, when $\tau$ is increased to 0.9, the visual result is already very similar to those obtained with neighboring high-threshold settings, suggesting that further increasing the threshold does not provide a noticeable improvement in artifact purification.

However, setting $\tau$ to 1.0 is not used as a valid experimental setting in our implementation. Such an excessively strict threshold would retain only points with nearly perfect semantic consistency across valid projections, leaving only a very small number of points after purification. Consequently, $\tau=1.0$ is regarded as a degenerate case rather than a meaningful purification setting.

Based on these observations, $\tau=0.9$ is adopted as the default setting. This value corresponds to a strict multi-view semantic consistency criterion: a point is retained only when it is classified as belonging to the artifact in at least 90\% of its valid projections. This setting effectively suppresses residual background points and boundary artifacts while preserving the main artifact structure. Therefore, $\tau=0.9$ provides a conservative but stable purification setting for generating artifact-only point clouds before registration.

\begin{table}[!htbp]
\centering
\caption{Effect of percentile intervals on thickness-axis extent estimation and rotation hypothesis selection.}
\label{tab:percentile_sensitivity}
\begin{tabular}{ccccc}
\toprule
\textbf{Object} & \textbf{Percentile interval} & \textbf{Front extent} & \textbf{Back extent} & \textbf{Selected hypothesis} \\
\midrule
\multirow{5}{*}{Artifact2}
& 1\%--99\%  & 0.30191 & 0.24756 & Identity hypothesis \\
& \textbf{2\%--98\%}  & \textbf{0.27836} & \textbf{0.23128} & \textbf{Identity hypothesis} \\
& 3\%--97\%  & 0.26240 & 0.22051 & Identity hypothesis \\
& 5\%--95\%  & 0.24273 & 0.20518 & Identity hypothesis \\
& 10\%--90\% & 0.20628 & 0.17403 & Identity hypothesis \\
\midrule
\multirow{5}{*}{Artifact4}
& 1\%--99\%  & 0.06878 & 0.13458 & Flip-Y hypothesis \\
& \textbf{2\%--98\%}  & \textbf{0.06330} & \textbf{0.04050} & \textbf{Flip-Z hypothesis} \\
& 3\%--97\%  & 0.05917 & 0.03443 & Flip-Z hypothesis \\
& 5\%--95\%  & 0.05044 & 0.02903 & Flip-Y hypothesis \\
& 10\%--90\% & 0.03712 & 0.02038 & Flip-Y hypothesis \\
\midrule
\multirow{5}{*}{Artifact6}
& 1\%--99\%  & 0.16997 & 0.21118 & Identity hypothesis \\
& \textbf{2\%--98\%}  & \textbf{0.14866} & \textbf{0.17846} & \textbf{Flip-Y hypothesis} \\
& 3\%--97\%  & 0.13794 & 0.16133 & Flip-Y hypothesis \\
& 5\%--95\%  & 0.12503 & 0.14470 & Flip-Y hypothesis \\
& 10\%--90\% & 0.10709 & 0.11877 & Flip-Y hypothesis \\
\bottomrule
\end{tabular}
\end{table}

\subsubsection{Sensitivity to Percentile Thresholds in Thickness Estimation}

The percentile thresholds used in thickness estimation determine the percentile-based bounds along the PCA-derived thickness axis. Since the resulting thickness-axis extent is further used to compute the thickness-derived offset and the ICP correspondence threshold during hypothesis evaluation, this parameter may affect both the estimated extent and the selected rotation hypothesis.

Table~\ref{tab:percentile_sensitivity} reports the influence of different percentile intervals on representative objects. Artifacts 2, 4, and 6 were selected because they cover a broad range of thickness scales and geometric conditions in the dataset. Artifact 4 represents a typical thin-sheet case, whereas Artifacts 2 and 6 are larger and relatively thicker fragments. This selection allows us to evaluate whether the percentile interval remains robust for both normal thin-sheet artifacts and fragments with larger thickness-axis extents. The results show that, as the percentile interval becomes narrower, the front-side and back-side thickness-axis extents generally decrease. This indicates that the percentile interval directly controls how much of the thickness-axis distribution is retained for thickness estimation.

When the interval is too wide, sparse extreme points caused by reconstruction noise or boundary artifacts may still be included. In this case, the thickness-axis extent can be overestimated, which may further affect the thickness-derived offset and the ICP correspondence threshold. As a result, the subsequent hypothesis evaluation may select a different rotation hypothesis.

When the interval is too narrow, more points near the two ends of the thickness-axis distribution are removed. Although this can suppress sparse extreme points, it may also discard valid geometric variation along the thickness direction. This can lead to an underestimated thickness-axis extent and may also change the selected rotation hypothesis.

The results in Table~\ref{tab:percentile_sensitivity} show that the adopted 2\%--98\% interval provides a balanced setting. It removes sparse extreme points more effectively than overly wide intervals, while avoiding the excessive trimming caused by overly narrow intervals. Therefore, the 2\%--98\% interval is used as the default percentile interval for robust thickness estimation in the proposed pipeline.

\begin{table*}[!htbp]
\centering
\caption{Sensitivity of fitness evaluation to different $\gamma$ values on representative objects.}
\label{tab:gamma_sensitivity}

\begin{tabular}{ccccccc}
\toprule
\textbf{Object} & \textbf{$\gamma$} & \textbf{Identity} & \textbf{Flip-X} & \textbf{Flip-Y} & \textbf{Flip-Z} & \textbf{Selected hypothesis} \\
\midrule
\multirow{6}{*}{Artifact4}
& 0.5 & 0.76839 & 0.84200 & \textbf{0.89447} & 0.81110 & Flip-Y hypothesis \\
& 1.0 & 0.91305 & 0.92541 & \textbf{0.97597} & 0.95562 & Flip-Y hypothesis \\
& 1.5 & 0.94665 & 0.94321 & \textbf{0.98011} & 0.97681 & Flip-Y hypothesis \\
& 2.0 & 0.95818 & 0.95464 & 0.99229 & \textbf{0.99823} & Flip-Z hypothesis \\
& 2.5 & 0.97021 & 0.96180 & 0.99949 & \textbf{1.00000} & Flip-Z hypothesis \\
& 3.0 & 0.98178 & 0.97244 & \textbf{1.00000} & \textbf{1.00000} & Ambiguous \\
\midrule
\multirow{6}{*}{Artifact6}
& 0.5 & 0.89332 & 0.90663 & \textbf{0.96612} & 0.88177 & Flip-Y hypothesis \\
& 1.0 & 0.98077 & 0.95568 & \textbf{0.98939} & 0.96194 & Flip-Y hypothesis \\
& 1.5 & 0.99745 & 0.98316 & \textbf{1.00000} & 0.98573 & Flip-Y hypothesis \\
& 2.0 & 0.99998 & 0.99765 & \textbf{1.00000} & 0.99515 & Flip-Y hypothesis \\
& 2.5 & \textbf{1.00000} & \textbf{1.00000} & \textbf{1.00000} & 0.99914 & Ambiguous \\
& 3.0 & \textbf{1.00000} & \textbf{1.00000} & \textbf{1.00000} & \textbf{1.00000} & Ambiguous \\
\midrule
\multirow{6}{*}{Artifact7}
& 0.5 & 0.53114 & 0.87536 & \textbf{0.97764} & 0.58902 & Flip-Y hypothesis \\
& 1.0 & 0.77289 & 0.93829 & \textbf{0.98575} & 0.83786 & Flip-Y hypothesis \\
& 1.5 & 0.87849 & 0.96224 & \textbf{0.98895} & 0.88839 & Flip-Y hypothesis \\
& 2.0 & 0.94078 & 0.97790 & \textbf{0.99150} & 0.92091 & Flip-Y hypothesis \\
& 2.5 & 0.96627 & 0.98882 & \textbf{0.99265} & 0.95209 & Flip-Y hypothesis \\
& 3.0 & 0.98313 & 0.99278 & \textbf{0.99355} & 0.97528 & Flip-Y hypothesis \\
\bottomrule
\end{tabular}

\end{table*}

\subsubsection{Sensitivity to the $\gamma$ Parameter in Fitness Evaluation}

The parameter $\gamma$ is used in the fitness evaluation to define the thickness-scaled ICP correspondence threshold. Specifically, the maximum correspondence distance is defined as $d_{\max}=\gamma h$, where $h$ denotes the estimated thickness. Therefore, $\gamma$ controls the tolerance used to count inliers during rotation hypothesis evaluation. A smaller $\gamma$ leads to a stricter correspondence threshold, whereas a larger $\gamma$ produces a more permissive threshold.

To analyze the influence of this parameter, we evaluate $\gamma$ values from 0.5 to 3.0 with an interval of 0.5. For each value, all rotation hypotheses are refined using point-to-plane ICP, and the corresponding fitness scores are reported in Table~\ref{tab:gamma_sensitivity}. The hypothesis with the highest fitness score is selected. Objects 4, 6, and 7 were selected because they provide different thickness scales and hypothesis ambiguity levels. Object 4 represents a typical thin-sheet case, Object 6 represents a larger and relatively thicker fragment, and Object 7 provides a larger texture-rich artifact. This selection allows us to examine whether the default $\gamma$ value provides a balanced thickness-scaled fitness evaluation across different geometric conditions, avoiding both overly strict correspondence rejection and excessive fitness saturation. When two or more hypotheses obtain the same highest fitness score, the result is marked as Ambiguous, because the fitness evaluation cannot uniquely determine a single rotation hypothesis.

\begin{figure}[!htbp]
\centering
\includegraphics[width=0.9\linewidth]{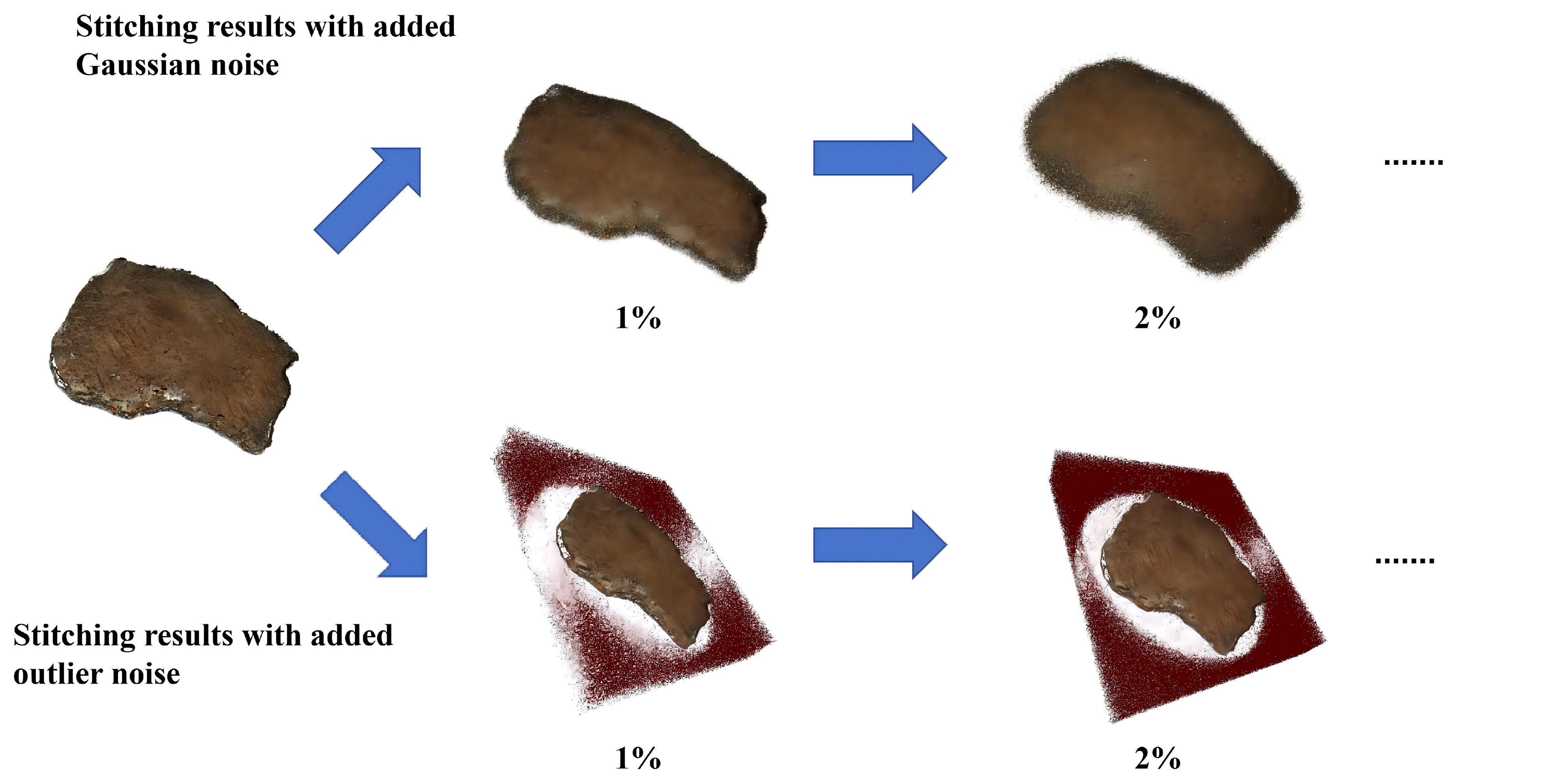}
\caption{Robustness evaluation under different noise perturbations. Two types of noise are introduced to the purified front--back point clouds, and the registration results are compared under increasing noise levels (0\%, 1\% and 2\%). The proposed method successfully aligns and stitches the point clouds even in the presence of noise, demonstrating stable registration performance.}
\label{fig:noise}
\Description{Two sequences illustrate registration robustness under increasing noise. The upper sequence adds Gaussian noise at zero, one, and two percent, producing an increasingly diffuse point distribution around the reconstructed surface. The lower sequence adds outlier noise at the same levels, shown as increasingly dense red points surrounding the artifact. In both sequences, the registered surfaces retain their overall alignment and recognizable shape as the noise level increases.}
\end{figure}

The results show that smaller $\gamma$ values produce lower fitness scores. This is because the correspondence threshold is more restrictive, and valid correspondences may be rejected under reconstruction noise, boundary artifacts, or residual alignment errors after ICP refinement. In this case, the fitness evaluation may become overly strict and sensitive to local reconstruction errors, which can affect the ranking of rotation hypotheses.

As $\gamma$ increases, the fitness scores generally become higher because more correspondences are counted as inliers. However, excessively large $\gamma$ values make the correspondence threshold too permissive. This reduces the discriminative ability of the fitness evaluation, since different rotation hypotheses may obtain identical or nearly identical high fitness scores. For example, in Table~\ref{tab:gamma_sensitivity}, Object 4 becomes Ambiguous when $\gamma=3.0$, and Object 6 becomes Ambiguous when $\gamma=2.5$ and $\gamma=3.0$. These cases indicate that overly large $\gamma$ values can lead to fitness saturation, making the selected hypothesis unreliable.

The adopted setting $\gamma=2.0$ provides a balanced choice in the proposed pipeline. Compared with smaller values, it provides sufficient tolerance for valid correspondences affected by reconstruction noise and residual misalignment. Compared with larger values, it avoids the stronger ambiguity caused by excessive relaxation of the correspondence threshold. Therefore, $\gamma=2.0$ is used as an empirical default setting for fitness evaluation in the following experiments.

\subsubsection{Robustness to Synthetic Noise}

To further evaluate the robustness of the proposed registration method under noisy conditions, additional experiments were conducted by introducing synthetic noise to the purified front--back point clouds. Specifically, random perturbations were added to the coordinates of both point clouds to simulate measurement errors that commonly occur in practical 3D reconstruction. Noise levels were considered, corresponding to 0\%, 1\% and 2\% of the scene scale. For each noise level, the disturbed point clouds were re-registered and stitched using the proposed method. The experimental results in Fig.~\ref{fig:noise} show that the method successfully aligns and merges the front and back point clouds across all tested noise levels. Even when the noise intensity increases from 0\% to 1\% and further to 2\%, the registration process remains stable and produces consistent alignment without structural misalignment or failure. These results demonstrate that the proposed method exhibits robustness to moderate geometric noise in the input point clouds.

\subsection{Ablation Study}

To analyze the contribution of each component in the proposed method, we conduct an ablation study by progressively removing key modules from the full pipeline. The proposed method integrates two major components: PCA-based canonical normalization with thickness estimation and geometry-constrained bidirectional registration with rotation hypothesis evaluation. To understand the individual impact of these components, we construct two degraded variants of the method and compare them with the full model.

First, we remove the thickness prior used in the registration stage. In this setting, the algorithm relies solely on geometric correspondence during point cloud alignment without enforcing the thickness-aware separation constraint described in Section 3.3. This variant evaluates the importance of incorporating physical thickness information for stabilizing the registration of thin sheet-like structures.

Second, we disable the rotation hypothesis enumeration step and perform registration using only a single initial orientation. Since PCA-based coordinate frames may exhibit sign ambiguity, the absence of hypothesis evaluation may cause the two surfaces to be incorrectly flipped or misaligned. This experiment evaluates the role of the rotation hypothesis search strategy in resolving orientation ambiguity between independently reconstructed surfaces.

\begin{figure}[!htbp]
\centering
\includegraphics[width=0.8\linewidth]{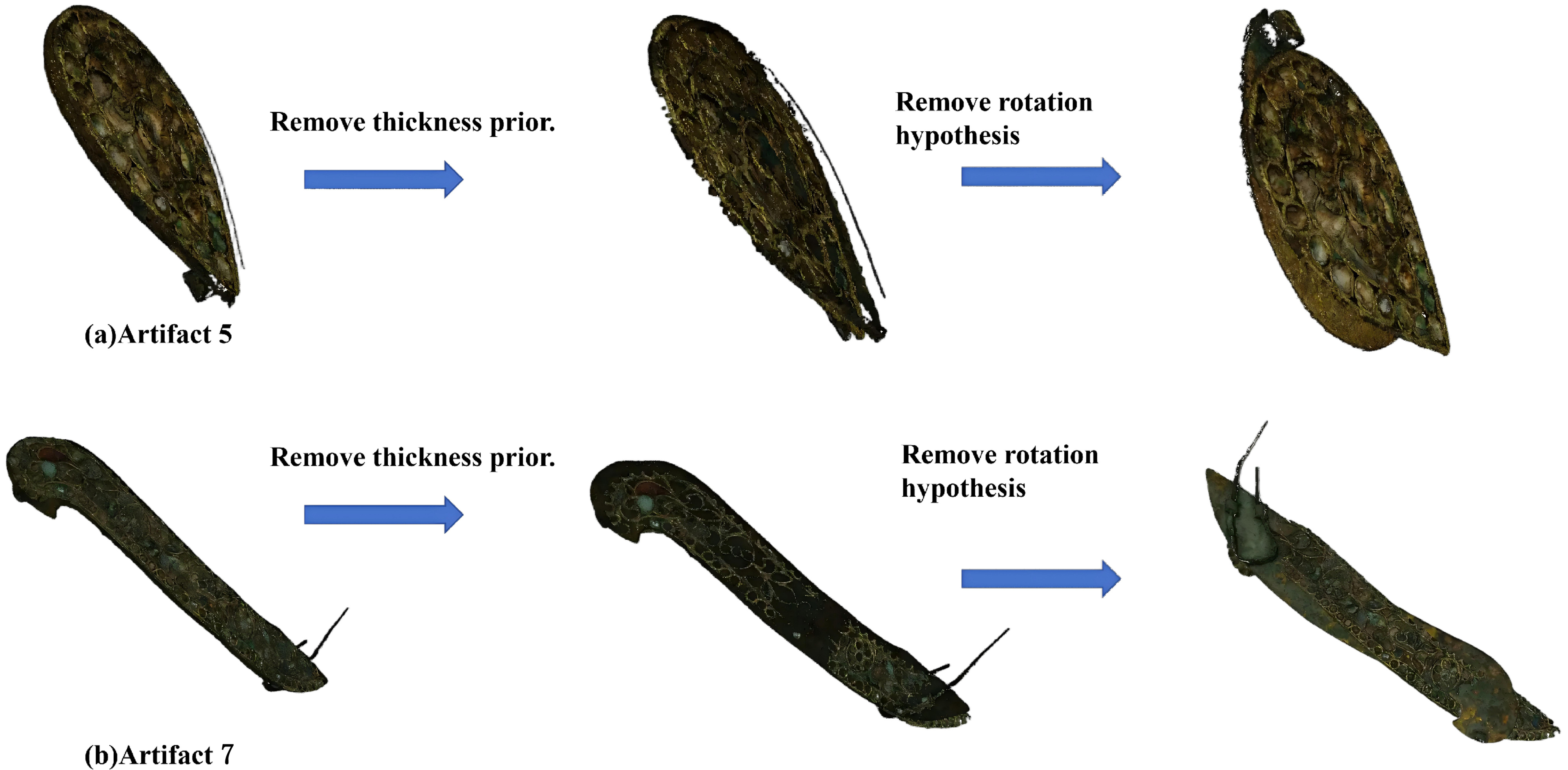}
\caption{Ablation study on two objects. From left to right: full model, without thickness prior, and without rotation hypothesis evaluation. Removing the thickness constraint leads to noticeable surface interpenetration, while further removing the rotation hypothesis causes severe misalignment between the reconstructed front and back surfaces.}
\label{fig:ablation}
\Description{Ablation results for two artifacts are arranged in two rows and three columns. Each row progresses from the full model on the left, to the result without the thickness prior in the center, and to the result without rotation-hypothesis evaluation on the right. Removing the thickness prior allows the two surfaces to intersect or collapse toward one another, while additionally removing hypothesis evaluation produces a visibly incorrect orientation and severe front-to-back misalignment.}
\end{figure}

As shown in Fig.~\ref{fig:ablation}, we select two representative objects to conduct the ablation experiments and visually analyze the influence of each component in the proposed pipeline.

After removing the thickness prior, we observe that the alignment becomes less constrained along the surface normal direction. As a result, the two reconstructed surfaces tend to interpenetrate, and noticeable surface piercing artifacts appear in the merged model. This phenomenon indicates that the thickness-aware constraint plays an important role in preventing unrealistic overlap between the front and back surfaces and helps stabilize the registration of thin structures. Furthermore, when the rotation hypothesis evaluation is also removed, the alignment quality deteriorates significantly. Without exploring multiple orientation candidates, the two surfaces may be incorrectly flipped or misoriented, leading to obvious mismatches between the front and back reconstructions. In several cases, the two sides of the object fail to align properly, suggesting that rotation hypothesis enumeration is important for mitigating orientation ambiguity introduced by PCA-based canonical frames.

The ablation results indicate that the thickness prior and rotation hypothesis evaluation are both important for stable front--back registration. Removing the thickness prior tends to produce physically implausible alignments, where the front and back surfaces collapse toward each other or partially interpenetrate. Disabling rotation hypothesis enumeration makes the registration more susceptible to orientation ambiguity, leading to incorrect front--back alignment. In contrast, the complete method preserves a more plausible thickness separation and yields visually more stable stitching results, demonstrating the effectiveness of combining PCA-based geometric normalization, rotation hypothesis evaluation, and thickness-aware registration.

\section{Conclusion and Future Work}

This paper presents a geometry-constrained bidirectional point cloud registration method for thin, sheet-like cultural heritage artifacts. It addresses the challenges of non-contact double-sided digitization, where the front and back surfaces are reconstructed independently with little or no geometric overlap.

The method combines semantic-guided purification, PCA-based canonical normalization, and thickness-aware registration. Semantic purification removes the supporting plane and background structures, while PCA normalization establishes a canonical coordinate system for geometric analysis and robust thickness estimation. Using this geometric prior, rotation hypothesis evaluation and point-to-plane ICP refinement are applied to resolve orientation ambiguity and reduce non-physical surface interpenetration during alignment.

Experimental results show that the proposed method achieves competitive or improved performance in most tested cases, particularly in projected area consistency and physically plausible front–back alignment. Compared with representative conventional and learning-based registration baselines, the proposed thickness-aware strategy reduces the risk of structural collapse when registering thin-sheet objects with limited geometric features, although certain baselines may obtain better image-space or silhouette metrics for some artifacts. Despite its effectiveness, the method may still fail for artifacts with strong geometric symmetry or weak boundary cues, where front–back orientation ambiguity can lead to inaccurate stitching. Further improvements are therefore needed to enhance robustness across diverse thin-sheet reconstruction scenarios.

Future work will explore several extensions. First, reference objects with known geometry, such as high-precision 3D-printed replicas, could provide more direct validation through explicit 3D geometric error metrics in addition to the current rendering- and projection-based metrics. Second, integrating the proposed registration strategy with neural rendering techniques such as 3D Gaussian Splatting may improve the visual fidelity of digital heritage models. Finally, the method could be extended to multi-fragment alignment for reconstructing fragmented cultural artifacts from independently scanned pieces.

\bibliographystyle{ACM-Reference-Format}
\bibliography{sample-base}

\end{document}